\documentclass[10pt,journal,compsoc]{IEEEtran}

\ifCLASSOPTIONcompsoc
  \usepackage[nocompress]{cite}
\else
  \usepackage{cite}
\fi

\usepackage{graphicx}
\usepackage{amsmath,amssymb} 
\usepackage{color}

\usepackage{multirow}

\usepackage{amsmath}

\usepackage{wrapfig}
\usepackage{float}
\usepackage{breqn}

\usepackage{subfigure}
\usepackage{booktabs}

\usepackage[misc]{ifsym}

\begin{document}
%
\title{Blind Motion Deblurring Super-Resolution: When Dynamic Spatio-Temporal Learning Meets Static Image Understanding}

\author{Wenjia Niu,
        Kaihao Zhang,
        Wenhan Luo,
        and Yiran Zhong
        \IEEEcompsocitemizethanks{ \IEEEcompsocthanksitem Wenjia Niu is with the School of electronic and information engineering, Hebei University of Technology, Tianjin, China. E-mail: \{niuwenjia9064@hotmail.com\}  \protect\\

\IEEEcompsocthanksitem Kaihao Zhang and Yiran Zhong are with the College of Engineering and Computer Science, Australian National University, Canberra, ACT, Australia. E-mail: \{kaihao.zhang@anu.edu.au; yiran.zhong@anu.edu.au\} \protect\\

\IEEEcompsocthanksitem Wenhan Luo is with Tencent, Shenzhen, China. E-mail: \{whluo.china@gmail.com\}  \protect\\

}

\thanks{Manuscript received April 19, 2005; revised August 26, 2015.}}

%
%

\markboth{Journal of \LaTeX\ Class Files,~Vol.~14, No.~8, August~2015}%
{Shell \MakeLowercase{\textit{et al.}}: Bare Advanced Demo of IEEEtran.cls for IEEE Computer Society Journals}
%



\IEEEtitleabstractindextext{%
\begin{abstract}

Single-image super-resolution (SR) and multi-frame SR are two ways to super resolve low-resolution images. Single-Image SR generally handles each image independently, but ignores the temporal information implied in continuing frames. Multi-frame SR is able to model the temporal dependency via capturing motion information. However, it relies on neighbouring frames which are not always available in the real world. Meanwhile, slight camera shake easily causes heavy motion blur on long-distance-shot low-resolution images. To address these problems, a \textbf{B}lind \textbf{M}otion \textbf{D}eblurring \textbf{S}uper-\textbf{R}eslution \textbf{Net}works, \textbf{BMDSRNet}, is proposed to learn dynamic spatio-temporal information from single static motion-blurred images. Motion-blurred images are the accumulation over time during the exposure of cameras, while the proposed BMDSRNet learns the reverse process and uses three-streams to learn Bidirectional spatio-temporal information based on well designed reconstruction loss functions to recover clean high-resolution images. Extensive experiments demonstrate that the proposed BMDSRNet outperforms recent state-of-the-art methods, and has the ability to simultaneously deal with image deblurring and SR.

\end{abstract}

}

\maketitle

\IEEEdisplaynontitleabstractindextext

%
\IEEEpeerreviewmaketitle

\section{Introduction}
\label{sec:introduction}
Super Resolution (SR) \cite{wang2020deep} has been an active topic for decades due to its utility in various applications. Its aims to improve the resolution of images given an input low-resolution image and output an image of high resolution. In most cases, low-resolution images also exhibit the artifact of blur. For example, capturing a fast-moving vehicle from a far distance produces an image of both low resolution and blur artifact. This paper focuses on super-resolving a low-resolution image with motion blur artifact.

Existing super resolution solutions approach the SR problem in both single-image \cite{niu2020single} and multi-frame ways \cite{kappeler2016video}. Solutions in the case of single image extract features in spatial domain only. These solutions can hardly work satisfactorily as they ignore the temporal information caused by the motion. Multi-frame super resolution is capable of using both spatial and temporal information contained in the multiple given frames, thus performs better than single-image super resolution. However, in our task, multiple neighbouring frames are not available.

As a fact, though we have only a single image, it does include rich temporal dynamics. The blurred single image can be considered as an overlay of a sequence of images shot in multiple time steps during the exposure time window (assuming we have a camera of higher shutter speed than the original camera capturing the given image) \cite{jin2018learning}. Once we are able to obtain the assumed sequence of multiple frames, multi-frame super resolution can spontaneously be carried out for better performance.

Inspired by this, this paper tries to learn the dynamic spatio-temporal information from a static motion-blurred low-resolution image. We decouple the problem into two sub problems, motion deblurring and (multi-frame) super resolution. For the first one, we aim to extract multiple clear frames from the given single motion-blurred image, thus extract the spatio-temporal information and solve the motion deblurring sub problem. For the second one, with the produced multiple frames, we conduct multi-frame super resolution by utilizing both temporal and spatial cues contained in the multiple low-resolution frames and produce a high resolution image free of blur defect. By doing so, we borrow the temporal dynamics enclosed in the static motion-blurred image and solve both the problem of blind motion blur and super resolution.

Specifically, for solving the motion deblurring problem, we propose a Blind Motion Deblurred Net (BMDNet) which is composed of convolutional layers and residual blocks. The BMDNet is successful in recovering a sequence of clear images from a given single motion-blurred image, disentangling the fused multiple images corresponding to finer-scale moments. As such, it solves the \texttt{static-to-dynamic} problem. Following the BMDNet there are parallel three streams within the BMDSRNet. To better utilize the bidirectional temporal cues contained in the sequence of frames. The first stream, ForNet, and the second stream, BackNet, process the recovered sequence from BMDNet in the forward and backward directions, respectively, with LSTM structure. The third stream is a CoreNet dedicated for the central frame. The three streams are learned with a well-designed reconstruction loss. The outputs of the three streams are fused by a FuNet to produce a high-resolution counterpart corresponding to the given low-resolution motion-blurred image. This thus tackles the \texttt{LR-to-HR} problem.

Our contributions are three-fold: 1) For the first time, we tackle the super resolution of single motion-blurred image as two sub-problems of motion deblurring and multiple frame super resolution. With this strategy, temporal cues can be utilized for the task of single image super resolution.
2) We propose an BMDSRNet to solve these two sub-problems. By learning bidirectional spatio-temporal dynamics with additional reconstruction loss, both these sub-problems are addressed. 3) Experimental results on public datasets demonstrate the superiority of the proposed method.

\section{Related Work}
\label{sec:related_work}

\subsection{Image Super-resolution}
Early solutions \cite{li2001new,zhang2006edge} to image super resolution address this problem with sampling based interpolation techniques. Natural image prior \cite{sun2008image}, neighbour embedding and sparse coding are employed to better predict finer textures by the subsequent works \cite{yang2012coupled,yang2010image}. 
Recently, deep learning achieves significant success in low-level vision tasks \cite{zheng2021t,zhang2021deep,zhang2021beyond,zhang2020beyond}, which also include image super-resolution \cite{dong2014learning, dong2015image,kim2016deeply,lim2017enhanced,lai2017deep,kim2016accurate,dong2016accelerating,ledig2017photo,wang2018esrgan,sajjadi2017enhancenet,zhang2018residual,zhang2018image,niu2020single,dai2019second,li2019feedback,zhang2020deep,zhang2018learning}.
In the deep learning, Dong \textit{et al.} \cite{dong2014learning} for the first time propose to solve the image SR problem using deep convolutional neural networks and the method surpasses the traditional methods. The success of residual structure in the recognition task inspires Kim \textit{et al.} to introduce the residual structure and thus train much deeper neural network for the image super-resolution task in \cite{kim2016accurate}. A deeply-recursive convolutional network (DRCN) is proposed in \cite{kim2016deeply}. A deep recursive layer is included in DRCN to improve the performance without new parameters. Huang \textit{et al.} employ bi-directional recurrent convolutional network to tackle the problem of video super-resolution in \cite{huang2015bidirectional}.
Generative Adversarial Networks (GAN) is introduced for image super resolution in \cite{Bulat_2018_ECCV}. A GAN based network is firstly trained to learn how to downgrade image resolution with unpaired data. The paired output of this network is used to train the desired image SR network. This method verifies its effectiveness in real-world images. The popular attention mechanism is introduced in a very deep residual channel attention networks (RCAN)\cite{Zhang_2018_ECCV} to improve the representation ability of CNNs and ease the difficulty of training deep networks for image SR task.

\begin{figure*}[!tb]
  \centering
\includegraphics[width=0.9\linewidth ]{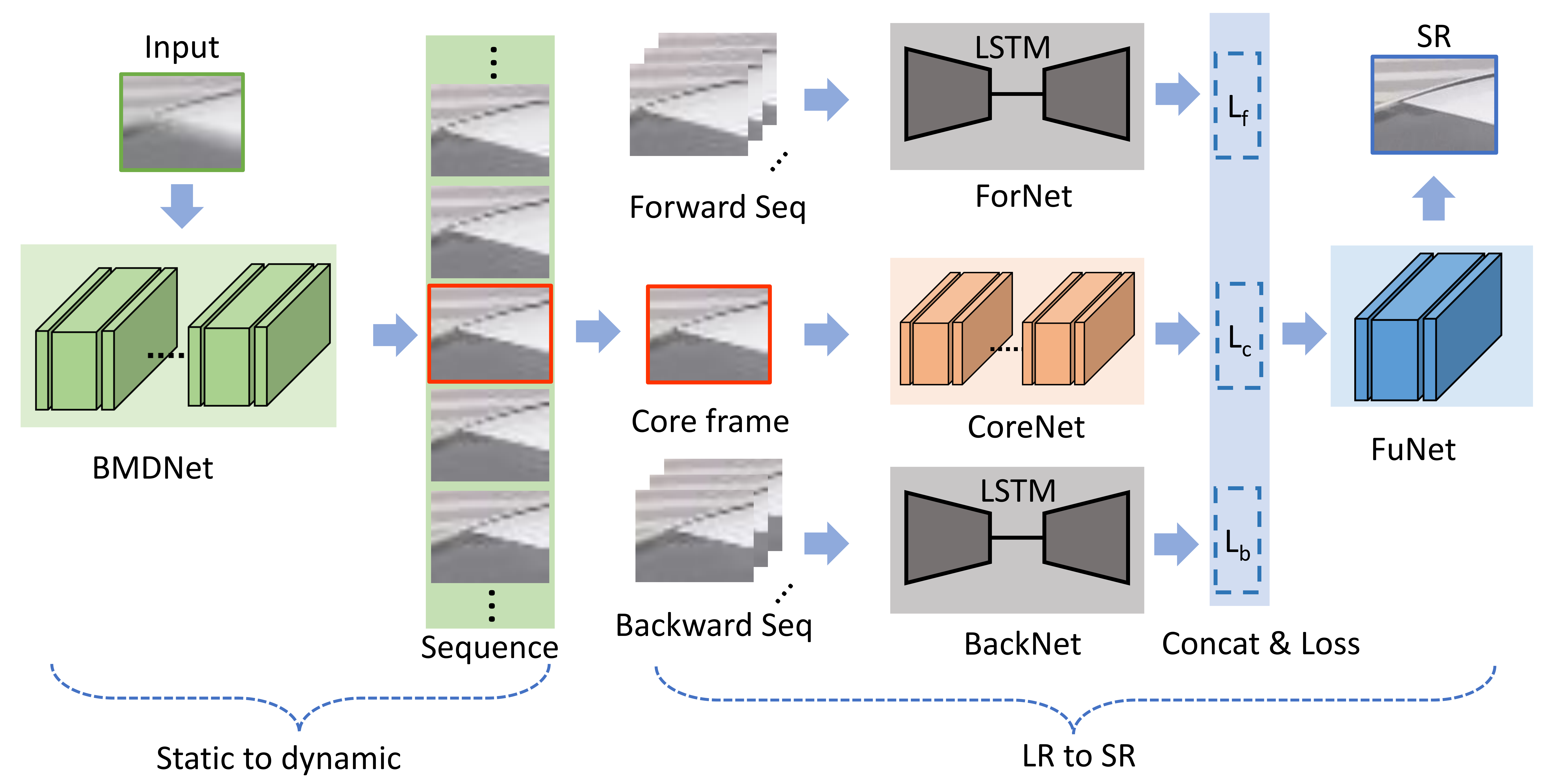}
\caption{{\bf The architecture of the motion deblurring super-resulution networks.} Static to dynamic: One Static low-solution motion blurred image is put into our model to remove unwanted blur and extract a video sequence. LR to SR: The generated video sequence is fed into three-stream networks, which includes ForNet, CoreNet and BackNet, to generate a high-solution image.}
\label{fig:architecture}
\end{figure*}

\begin{figure*}[t]
  \centering
  \subfigure[Decompose of blurred objects.]{
    \label{fig:motion_blur_a} 
    \includegraphics[width=0.6\linewidth]{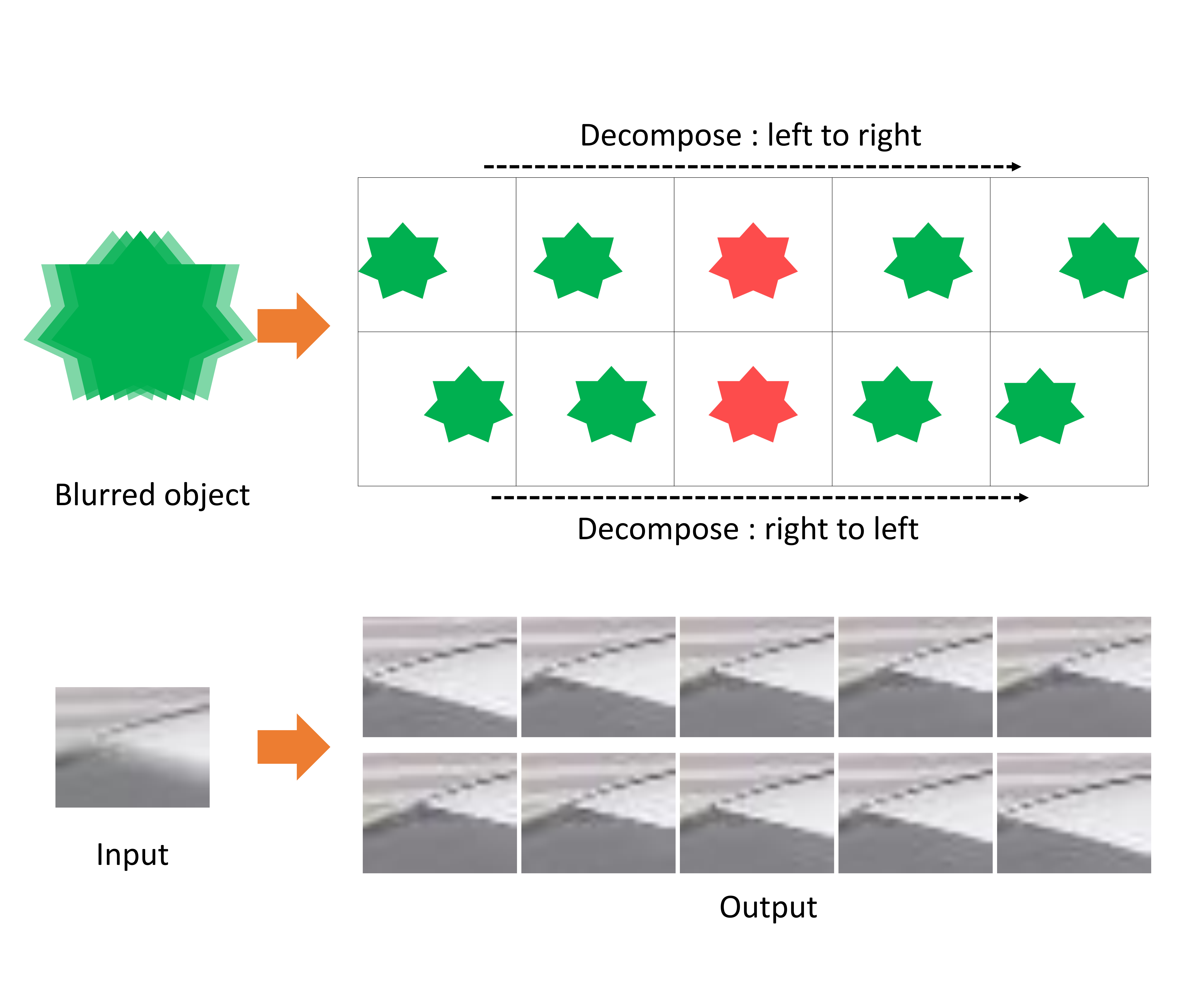}}
  \subfigure[Generated video sequences.]{
    \label{fig:motion_blur_b} 
    \includegraphics[width=0.6\linewidth]{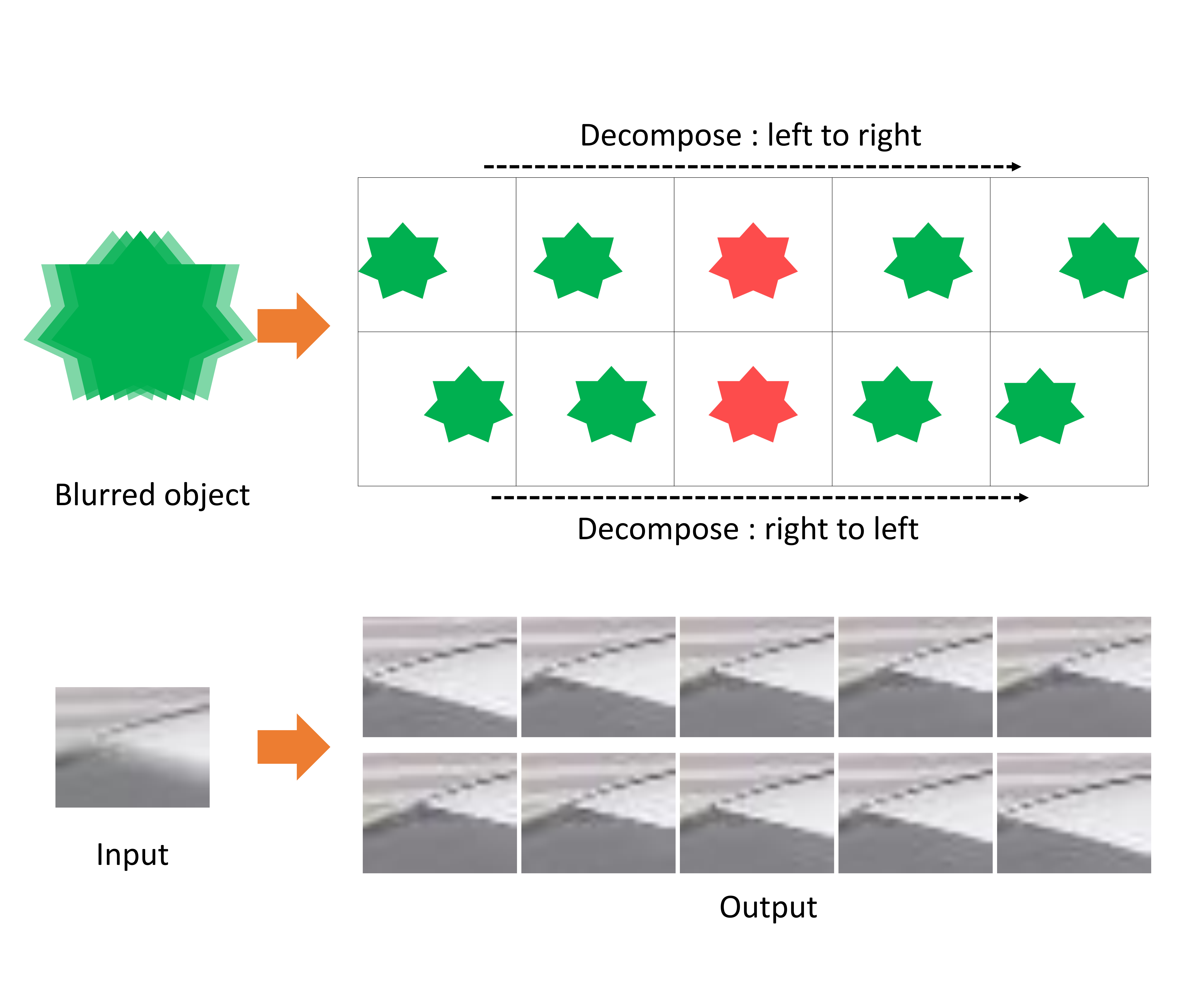}}
  \caption{{\bf The analyses of motion blurred images.} (a): Instant frames are accumulated over time to create a motion blurred images, which thus can be decomposed to different sequence. (b): Two different video sequences from a same motion blurred image.}
  \label{fig:motion_blur}
\end{figure*}

\begin{figure*}[!tb]
  \centering
\includegraphics[width=0.9\linewidth ]{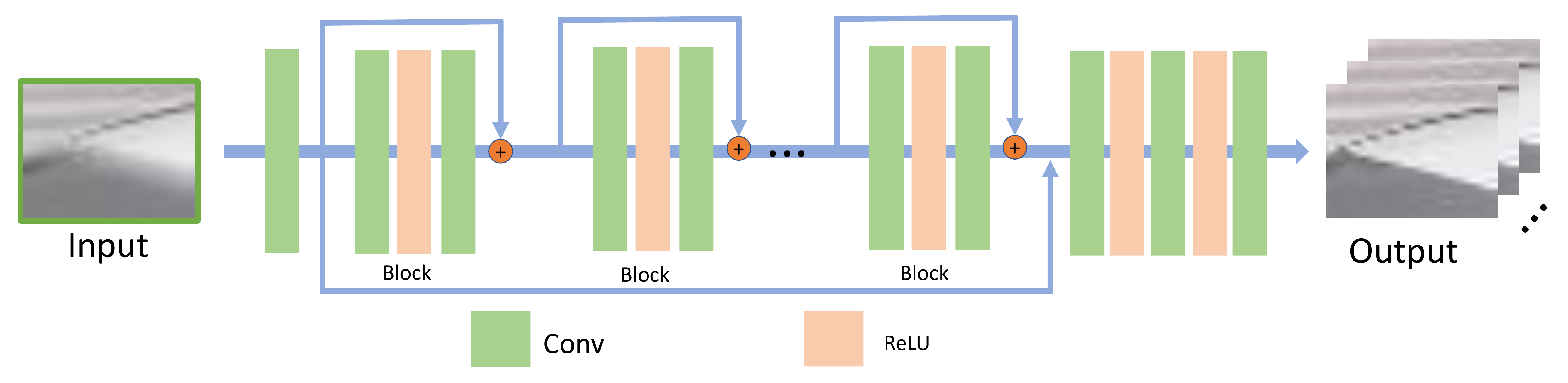}
\caption{{\bf The architecture of the BMDNet.} One blurred image is put into BMDNet to generate a video sequence.}
\label{fig:MDNet}
\end{figure*}

\begin{figure*}[!tb]
  \centering
\includegraphics[width=0.9\linewidth ]{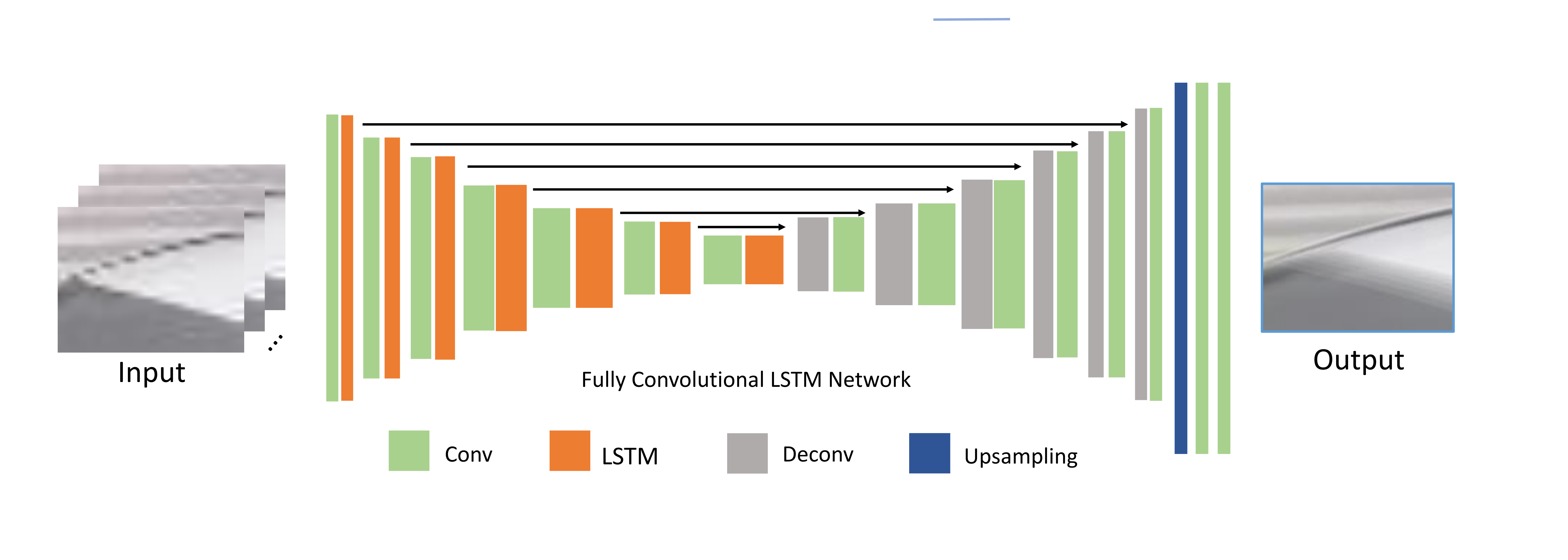}
\caption{{\bf The architecture of the ForNet and BackNet.} One video sequence is put into to generate a SR clean image. The input are the multi-frames generated by the BMDNet.}
\label{fig:ForNet_BackNet}
\end{figure*}

\subsection{Image Deblurring}
Image deblurring has been addressed by early methods based on priors or constraints \cite{chan1998total,shan2008high,krishnan2009fast}. These kind of solutions operates on multiple scales which are also time consuming. Recently, many deep deblurring methods are proposed to address the problem of image deblurring \cite{sun2015learning,chakrabarti2016neural,nah2017deep,nimisha2017blur,xu2017motion,zhang2018dynamic,tao2018scale,kupyn2018deblurgan,madam2018unsupervised,jin2018learning,ren2019neural,mustaniemi2019gyroscope,gao2019dynamic,lu2019unsupervised,aljadaany2019douglas,kupyn2019deblurgan,purohit2019region,zhang2020deblurring,suin2020spatially,kaufman2020deblurring,jiang2020learning} and video deblurring \cite{su2017deep,hyun2017online,aittala2018burst,zhang2018adversarial,chen2018reblur2deblur,nah2019recurrent,wang2019edvr,zhou2019spatio,pan2020cascaded,li2021arvo,zhang2020every}.
Schuler et al. \cite{schuler2015learning} propose a pioneer work of using deep learning for image deblurring. A two-stage architecture is developed and trained in an end-to-end manner for the blind image deblurring. Sun et al. \cite{sun2015learning} use CNN to estimate blur kernel and deblur images based on the estimated kernel. Targeted on non-blind deblurring, Xu \textit{et al.} \cite{xu2014deep} develop connection between deep neural networks and traditional optimization based approaches, propose a structure of two sub-modules, and achieve better performance.
Similar to the traditional methods before deep learning era \cite{shen2018deep,shen2019human,purohit2019bringing}, the multi-scale strategy is also employed by Nah \textit{et al.}~\cite{nah2017deep}. The sharp image is directly generated by a network without estimating the unknown kernel in \cite{chakrabarti2016neural}. CNN is also utilized along with RNN for image deblurring in \cite{zhang2018dynamic}. LSTM and CNNs are combined in a proposed SRN-DeblurNet to tackle image deblurring in a multi-scale manner by Tao \textit{et al.}~\cite{tao2018scale}. A nested skip connection structure is developed in \cite{gao2019dynamic}. In addition, there also exist some methods focusing on recover sharp high-resolution images from blurry low-resolution images. Zhang \textit{et al.} \cite{zhang2019deep2} propose a deep plug-and-play super-resolution network to handle LR image with arbitrary blur kernels. This method is a non-blind deblurring SR method, which relies on known blur kernels. For blind deblurring SR, the most related work is GFN \cite{zhang2018gated}. This method directly extracts the spatial information from a motion-blurred image to recover its sharp SR version, but ignores the temporal information implied in the motion-blurred image.

In this paper, we address the problem of blind motion deblurring super-resolution, which is a more difficult task than the individual problems of image deblurring and super-resolution. One reason is that the motion-blurred image includes spatio-temporal information, which is difficult to extract. In this paper, we propose a new method to address this problem via employing the “divide and conquer” scheme. Specially, to extract the spatio-temporal information, we first use a “from static to dynamic” network to generate a sequence of LR sharp images from an input motion blurred image. During the training stage, the input and output of the “from static to dynamic” network are blurry image and its corresponding sharp frames, respectively. Therefore, it has better ability to extract the information implied in the motion-blurred image. Then, we can use the restored sharp images to help the following network to finish the video super-resolution process and obtain the final results.

\section{Proposed Method}

\subsection{The Overall Architecture of BMDSRNet}
Fig. \ref{fig:architecture} shows the overview architecture of BMDSRNet. Primarily, it is composed of two parts, aiming to solve the ``static-to-dynamic" and ``LR-to-HR" problems individually. The first part is a designated Blind Motion Deblurring Net (BMDNet), which takes a given low resolution image with motion blur as input and outputs a sequence of motion deblurred images, corresponding to the multiple clear images during the exposure time period. This part does not address the resolution issue. The second part specifically focuses on the super-resolution aspect. This part consists of three parallel streams, a ForNet, a CoreNet, and a BackNet. The ForNet processes the sequence of images from BMDNet in a forward direction. On the contrary, the BackNet learns the backward temporal information by processing the reverse order of the sequence. Both of ForNet and BackNet are of the LSTM structure and share weights during the training stage. The CoreNet operates on the central frame as it is the most important one among the sequence. The outputs of these three streams are concatenated and fused by a FuNet aiming at recovering a high resolution image corresponding to the given low resolution image, free of motion blur.

\subsection{From Static to Dynamic}
The first part in the proposed BMDSRNet aims to extract a set of continuous frames from a given motion-blurred image. 
The intuition behind is that, a motion blurred image can be considered as an accumulation of multiple instant frames. By decomposing the blurred image into multiple clear images with BMDNet (as Fig. \ref{fig:motion_blur} shows), we can transform a \emph{static} motion-blurred image into a sequence of \emph{dynamic} frames. This strategy benefits our ultimate goal in two aspects. Firstly, this kind of decomposition solves the deblurring problem. Secondly, by transforming a single image into multiple continuous frames, the difficulty of super-resolution is eased as we can subsequently utilize the spatio-temporal cues contained in the multiple resulted frames, as illustrated in \cite{huang2015bidirectional}.

To accomplish this \emph{``static-to-dynamic"} task, we develop a neural network called BMDNet. Fig. \ref{fig:MDNet} presents the structure of BMDNet. The input is a single motion-blurred image and the output is seven continuous clear frames. There are sequentially two convolutional layers of kernel size $3\times3$, nine residual blocks \cite{he2016deep} and three convolutional layers of kernel size $3\times3$. Through the last convolutional layer, the output feature channels turn 21, corresponding to seven frames. Table \ref{table:MDNet} illustrates the detailed configuration of BMDNet.

Let the input of BMDNet be denoted as $X_{blurry}$, and the output as $\{Y_{out}^i, i=1, 2, ..., 7\}$. In our practice, we use seven consecutive clear frames $\{Y_{sharp}^i, i=1, 2, ..., 7\}$ to synthesize the motion-blurred image. Thus, the used seven clear frames serve as the ground truth frames. The cost function of training the BMDNet is composed of a term of reconstruction loss $\mathcal{L}^c_{S2D}$ regarding the central frame and terms of pair-wise content loss $\mathcal{L}^p_{S2D}$ regarding the other six frames. The ``S2D" here indicates ``static-to-dynamic".

The reconstruction content loss of the central frame is

\begin{equation}
\mathcal{L}^c_{S2D} = \left\| Y_{sharp}^4 -  Y_{out}^4 \right\| + \left\|\Phi(Y_{sharp}^4)-\Phi(Y_{out}^4)\right\|\, ,
\end{equation}
where the $\Phi$ extracts features using the last convolutional layer of VGG19 network \cite{simonyan2014very}. This loss term measures both the pixel-wise and perceptual level error between the recovered frame and the ground truth.

Given a set of continuous images, changing the order of images would not change the accumulation of multiple frames. This means there could be multiple possible solutions to the reverse process, \emph{i.e.} extracting multiple frames from a given image. Fig. \ref{fig:motion_blur_a} and \ref{fig:motion_blur_b} respectively show the concept and an example of decomposing a single image into possible two sets of sequential images. 
This means it is not appropriate to use image-wise content loss like the central frame for the other six images. Thus, following \cite{Jin_2018_CVPR}, we also employ a pair-wise content loss to ensure the output sequence frames are reasonable, formulated as,
\begin{equation}
\small
\begin{split}
& {\mathcal{L}_{S2D}^{p}} = \sum\nolimits_{i = 1}^3 \left(  {[Y_{sharp}^{i}, Y_{sharp}^{8 - i}]_ + } - {[Y_{out}^i,Y_{out}^{8-i}]_ + } \right) \\
& \ \ \ \ \ \ \ \ \ \ + \left( {[Y_{sharp}^{i},Y_{sharp}^{8-i}]_ - } - {[Y_{out}^i, I_{out}^{8-i}]_ - } \right) ,
\end{split}
\end{equation}

where ${[\cdot, \cdot]_+}$ and ${[\cdot, \cdot]_-}$ denote the pixel-wise summation and subtraction between a pair of images, respectively.

Notably, it is not the first time in the community that a neural network is proposed to extract multiple frames from a single static image. \cite{Jin_2018_CVPR} is a pioneer work. However, our BMDNet is distinctly different from the method in \cite{Jin_2018_CVPR}. Our method uses only a single model to carry out the task of extracting multiple frames, but \cite{Jin_2018_CVPR} requires multiple models for the task (as many as the number of output frames), which is not efficient in practice.

In the real world, it is difficult to determine if the input image is “LR” or “HR”. This is because the LR and HR are relative concepts. For example, the 4K images are high-resolution compared to the 2K images, but are low-resolution compared to the 8K versions. The “From static to dynamic” network is proposed to extract a sequence of sharp images from a motion-blurred image. Therefore, we mainly ensure that the input and ground truth images are the motion-blurred images and their seven neighboring sharp images during the training stage.

\begin{table}[!tb]
  \centering
  \small
  \caption{Configurations of the proposed BMDNet. It is composed of two convolutional layers (L1 and L2), nine residual blocks and three additional convolutional layers (L3, L4 and L5). Each residual block contains two convolutional layers. The output size is the same to input.}
    \begin{tabular}{c c c c}
    \toprule
    layers & Kernel size & output channels & operations \\
    \midrule
     L1     & $3 \times 3 $  & 64 & -  \\
     L2       & $3 \times 3 $  & 64 & ReLU \\ 
     \midrule
     B1-B8     & $3 \times 3 $  & 64 & ReLU \\
     B9    & $3 \times 3 $  & 64 & ReLU \\
     \midrule
     L3    & $3 \times 3 $  & 64 & ReLU \\
     L4    & $3 \times 3 $  & 64 & ReLU \\
     L5    & $3 \times 3 $  & 21 & - \\
    \bottomrule
    \end{tabular}
  \label{table:MDNet}
\end{table}

\begin{table}
  \centering
  \small
  \caption{Configurations of the proposed ForNet and BackNet. It is composed of convolutional, ConvLSTM and deconvolutional layers. The output size is $N$ times larger than input.}
    \begin{tabular}{c c c}
    \toprule
    layers & Kernel size & output channels  \\
    \midrule
     Conv+ConvLSTM     & $3 \times 3 $  & 32  \\
     Conv+ConvLSTM     & $3 \times 3 $  & 64  \\
     Conv+ConvLSTM     & $3 \times 3 $  & 128  \\
     Conv+ConvLSTM     & $3 \times 3 $  & 256  \\
     Conv+ConvLSTM     & $3 \times 3 $  & 256  \\
     Conv+ConvLSTM     & $3 \times 3 $  & 256  \\
     Conv+ConvLSTM     & $3 \times 3 $  & 512  \\
     \midrule
     DeConv+Concat+Conv     & $3 \times 3 $  & 256  \\
     DeConv+Concat+Conv     & $3 \times 3 $  & 128  \\
     DeConv+Concat+Conv     & $3 \times 3 $  & 128  \\
     DeConv+Concat+Conv     & $3 \times 3 $  & 128  \\
     DeConv+Concat+Conv     & $3 \times 3 $  & 64  \\
     DeConv+Concat+Conv     & $3 \times 3 $  & 32  \\
     DeConv     & $3 \times 3 $  & 32  \\
     \midrule
     Upsampling+Conv        & $3 \times 3 $  & 32 \\
     Conv                   & $3 \times 3 $  & 3 \\
    \bottomrule
    \end{tabular}
  \label{table:FBNet}
\end{table}

\begin{figure*}[!tb]
  \centering
\includegraphics[width=0.9\linewidth ]{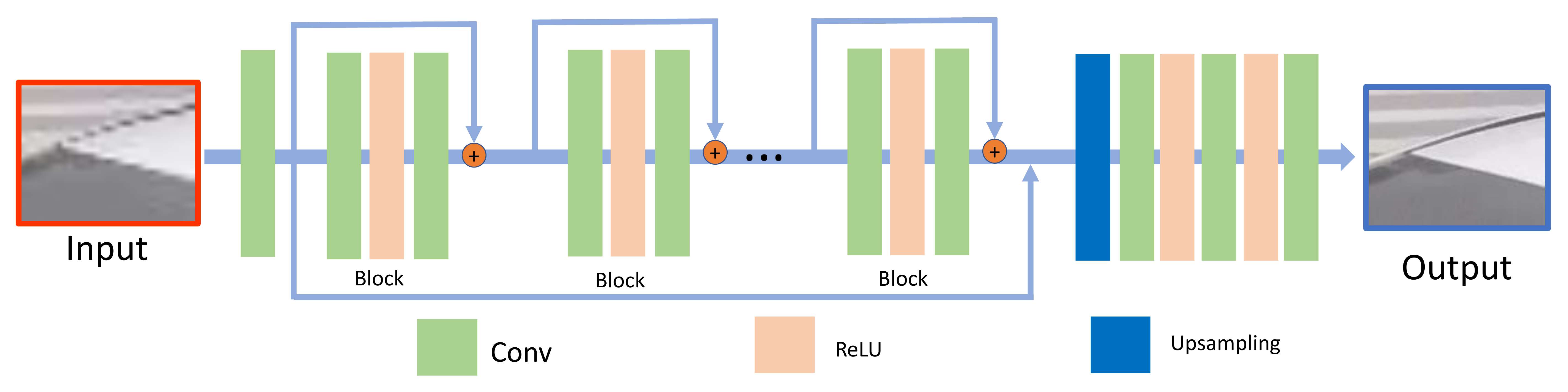}
\caption{{\bf The architecture of the CoreNet.} One low-level image is put into to generate a super-resolution image. Then input is the center frame generated by BMDNet.}
\label{fig:CoreNet}
\end{figure*}

\begin{figure*}[!tb]
  \centering
\includegraphics[width=0.9\linewidth ]{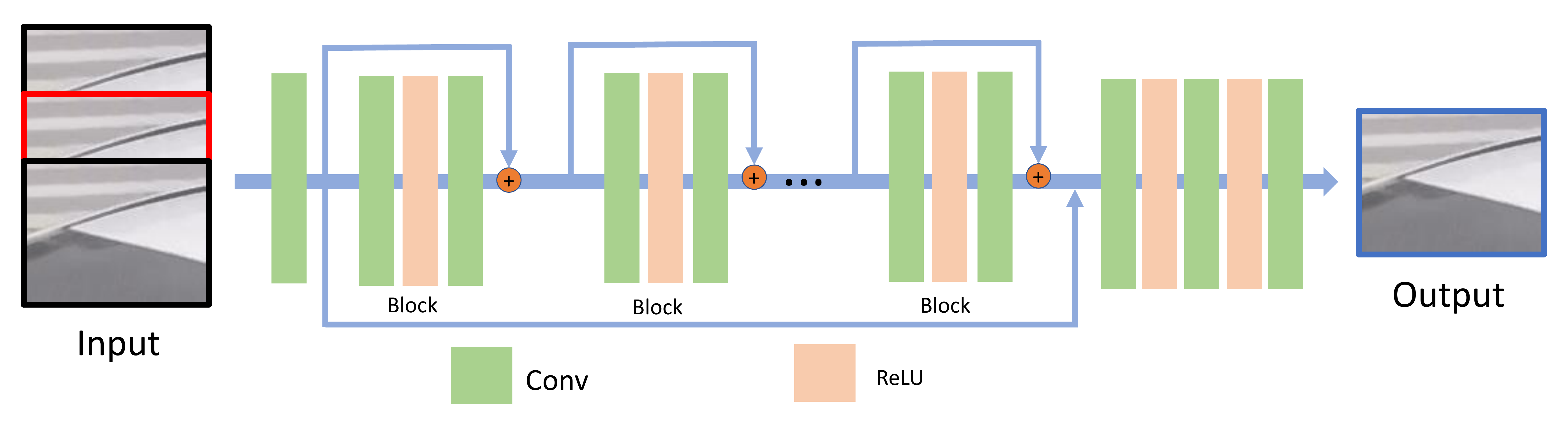}
\caption{{\bf The architecture of the FuNet.} Three intermediate results generated by preceding networks are utilized to recover the final finer super-resolution images. Then input are three images, which are the output of ForNet, CoreNet and BackNet, respectively.}
\label{fig:FuNet}
\end{figure*}

\subsection{From Low-resolution to High-resolution}
The first part of BMDSRNet, \emph{i.e.} BMDNet, has decomposed a single low resolution into seven low resolution frames free of blur artifact. These multiple continuous frames benefit the multi-frame super resolution task, accomplished by the second part.

There are three streams in the second part of BMDSRNet, \emph{i.e.}, ForNet, CoreNet and BackNet. As mentioned before, there could be multiple reasonable solutions to the ``static-to-dynamic" sub-problem.
Fig. \ref{fig:motion_blur} shows an example of two possible solutions of different directions, which both can result in the blurred image. This inspires us to learn the temporal dynamics in both the forward and backward directions, to cover the variety. To this end, the ForNet and BackNet are designed for the purpose of learning both the forward and backward motion.

ForNet differs from BackNet in terms of the input. We reverse the order of input frames of ForNet as the input to the BackNet. 
Both of them are networks of fully convolutional layers with LSTM structure. As Fig. \ref{fig:ForNet_BackNet} shows, taking ForNet as an example, it consists of several layers of convolution and ConvLSTM, followed by several layers of deconvolution, convolution and upsampling. The size of all the kernels is $3\times 3$. Table \ref{table:FBNet} represents the configurations of ForNet and BackNet. 

The CoreNet is specifically for the central frame, as the center frame is the most important one in the sequence. Similarly, it is composed of a sequence of convolutional layers, residual blocks and convolutional layers, as shown in Fig. \ref{fig:CoreNet}. The detailed configurations of CoreNet is given in Table \ref{table:CoreNet}.

In order to push the intermediate output become the sharp high-resolution images, we train out ForNet, BackNet and CoreNet based on MSE criterion. The loss function can be formulated as:

\begin{equation}
{\mathcal{L}_{content}} = \frac{1}{{WH}}\sum\limits_{x = 1}^W {\sum\limits_{y = 1}^H {{{(I_{x,y}^{sharp} - G(I^{blurry})_{x,y})}^2}} }\, ,
\end{equation}
where $W$ and $H$ are the width and height of a frame, $I_{x,y}^{sharp}$ is the value of high-resolution images at location $\left(x,y\right)$, and $G(I^{blurry})_{x,y}$ corresponds to the value of super-resolution images which are recovered from ForNet and BackNet.

Finally, we develop a FuNet to fuse three different intermediate results and generate the final finer high-resolution images. FuNet has a similar structure as CoreNet, but with two primary differences. Firstly, it is a smaller structure, which has only three blocks. The reason is that the input to FuNet is already high-quality super-resolution images generated by the preceding sub-networks. The second difference is that FuNet does not have the upsampling layer. Fig. \ref{fig:FuNet} shows the architecture of FuNet.

\begin{table}
  \centering
  \small
  \caption{Configurations of the proposed CoreNet. It is composed of two convolutional layers (L1 and L2), 9 residual blocks and three additional convolutional layers (L3, L4 and L5). Each residual block contains two convolutional layers. The output size is $N$ times larger than input.}
    \begin{tabular}{c c c c}
    \toprule
    layers & Kernel size & output channels & operations \\
    \midrule
     L1     & $3 \times 3 $  & 64 & -  \\
     L2       & $3 \times 3 $  & 64 & ReLU \\ 
     \midrule
     B1-B8     & $3 \times 3 $  & 64 & ReLU \\
     B9    & $3 \times 3 $  & 64 & ReLU+Upsampling \\
     \midrule
     L3    & $3 \times 3 $  & 64 & ReLU \\
     L4    & $3 \times 3 $  & 64 & ReLU \\
     L5    & $3 \times 3 $  & 3 & - \\
    \bottomrule
    \end{tabular}
  \label{table:CoreNet}
\end{table}

\section{Experiments}

\subsection{Motion-Blurred LR Dataset}

One of the most popular motion-blurred datasets is the GOPRO dataset \cite{nah2017deep}. Using a high-speed camera, Nah \textit{et al.} capture $33$ videos ($22$ for training and $11$ for testing respectively), and then synthesize $3,214$ pairs of blurry image and sharp image for training and testing. This synthesized dataset cannot be utilized to train our model directly because we need a blurry image and its corresponding seven sharp images. to learn the process of \textit{``static-to-dynamic"}. In order to address these problem, we re-synthesize a Motion-Blurred LR dataset based on the GOPRO dataset. We firstly extract all the frames from the $33$ provided videos. Then we synthesize blurry images via averaging the seven neighbouring sharp images. By doing this, a motion-blurred image corresponds to seven sharp images and thus we can train our BMDNet to learn the process of \textit{``static-to-dynamic"}. In order to model the stage of  \textit{``from low-resolution to high-resolution''}, we further down-sample the images to generate low-resolution counterparts. The training and testing sets are generated based on $22$ and $11$ videos, respectively, which is same as the original GOPRO dataset.

\begin{figure*}[!tb]
  \centering
  \subfigure[]{
    \label{idea:a}
    \includegraphics[width=0.18\linewidth ]{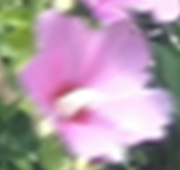}}
  \subfigure[]{
    \label{idea:b}
    \includegraphics[width=0.18\linewidth ]{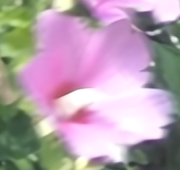}}
    \subfigure[]{
    \label{idea:c}
    \includegraphics[width=0.18\linewidth ]{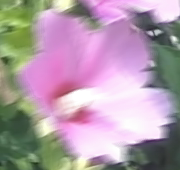}}
  \subfigure[]{
    \label{idea:d}
    \includegraphics[width=0.18\linewidth ]{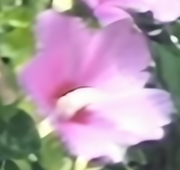}}
    \subfigure[]{
    \label{idea:e}
    \includegraphics[width=0.18\linewidth ]{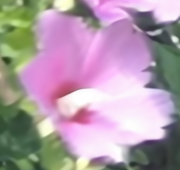}}
\caption{{\bf Examples of blind motion deblurring super-resolution.} From left to right are the input image, deblurring SR results from SRNet, BMDSRNet(C), BMDSRNet(F+C) and BMDSRNet(F+C+B), respectively. Zoom in the figure for better visibility.}
\label{figure_ablation_study}
\end{figure*}

\begin{table}[!tb]
  \centering 
   \small
    \caption{Ablation study for scale factor 2, 3 and 4 on the Motion-Blurred LR dataset in terms of PSNR and SSIM.}
    \setlength\tabcolsep{5.0pt}
    \begin{tabular}{l | c | c c }
    \toprule
    Scale & Methods &  PSNR & SSIM \\
    \hline
    $\times$ 2 & SRNet & 30.88 & 0.9437\\
    $\times$ 2 & BMDSRNet(C) & 31.28 & 0.9457 \\
    $\times$ 2 & BMDSRNet(F+C) & 31.45 & 0.9479  \\
    $\times$ 2 & BMDSRNet(F+C+B) & 31.62 & 0.9483 \\
    \hline
    $\times$ 3 & SRNet & 29.49 & 0.9256 \\
    $\times$ 3 & BMDSRNet(C) & 29.95 & 0.9302 \\
    $\times$ 3 & BMDSRNet(F+C) & 30.21 & 0.9337  \\
    $\times$ 3 & BMDSRNet(F+C+B) & 30.44 & 0.9356 \\
    \hline
    $\times$ 4 & SRNet & 28.06 & 0.8997\\
    $\times$ 4 & BMDSRNet(C) & 28.43 & 0.9068 \\
    $\times$ 4 & BMDSRNet(F+C) & 28.62 & 0.9113  \\
    $\times$ 4 & BMDSRNet(F+C+B) & 28.78 & 0.9132 \\
    \bottomrule
    \end{tabular}%
    \label{table_ablation}
\end{table}%

\begin{figure*}[!tb]
  \centering
  \subfigure[]{
    \label{idea:b}
    \includegraphics[width=0.19\linewidth ]{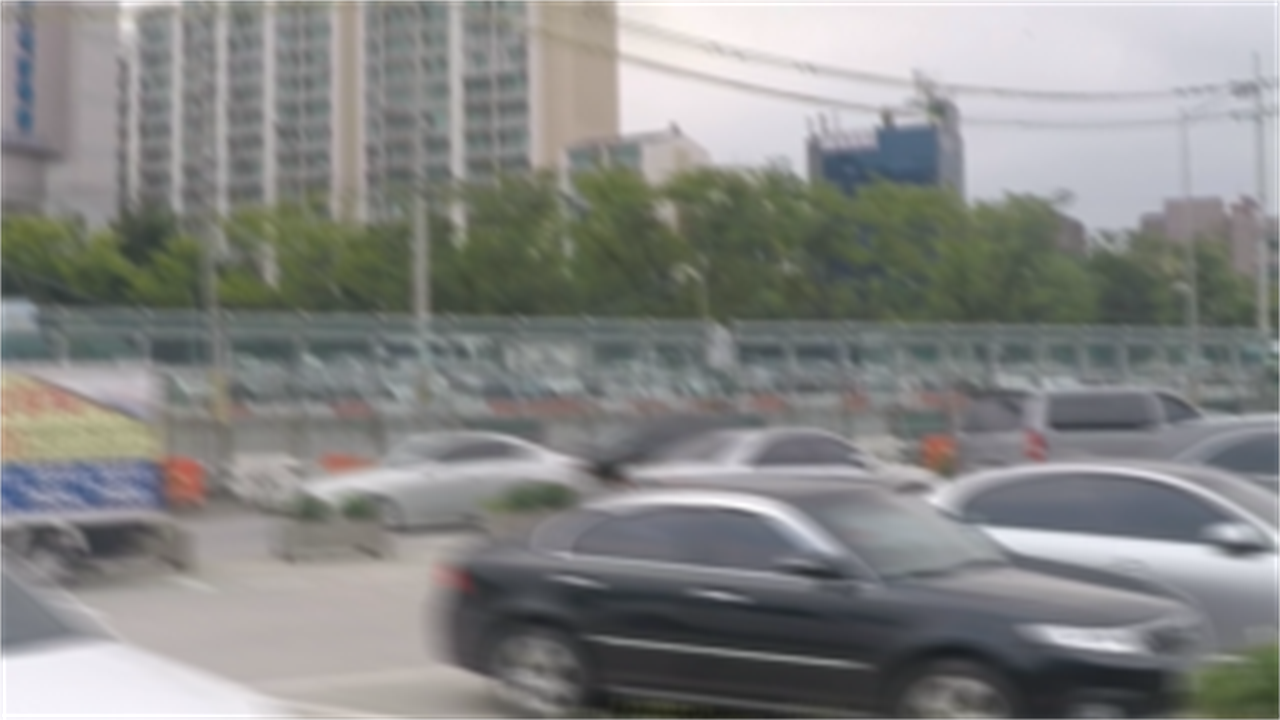}}
  \subfigure[]{
    \label{idea:d}
    \includegraphics[width=0.19\linewidth ]{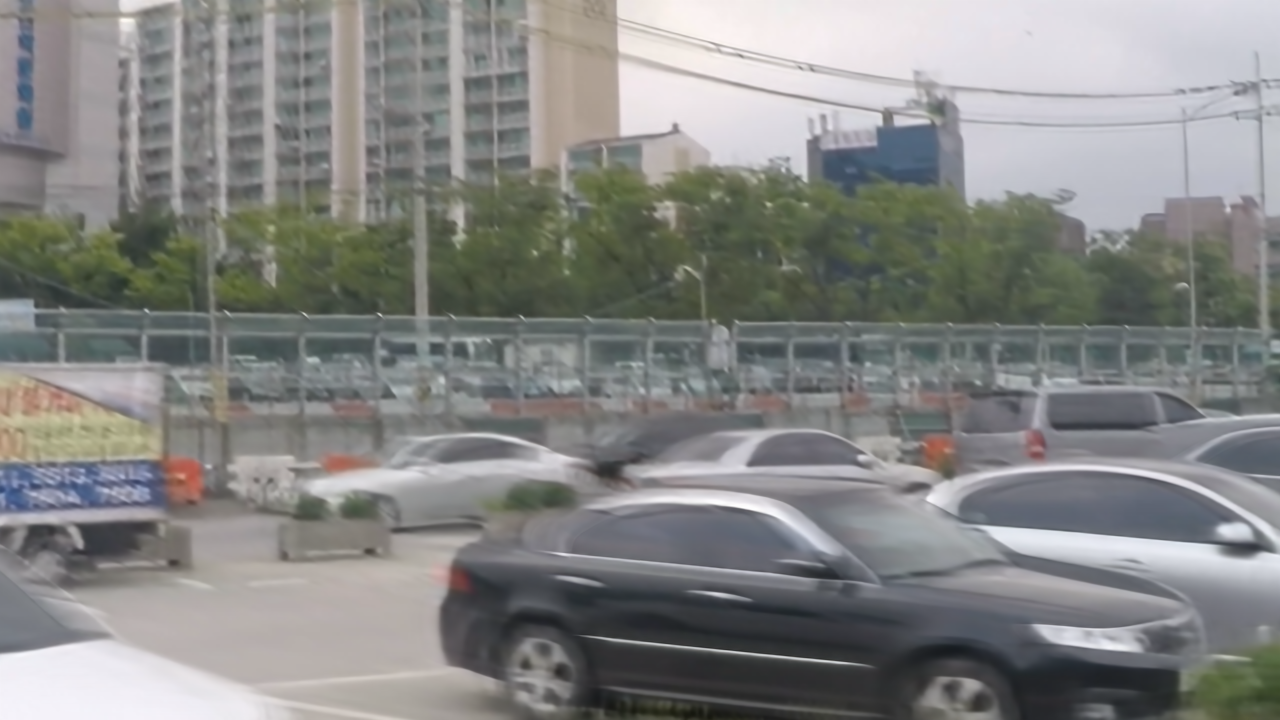}}
    \subfigure[]{
    \label{idea:f}
    \includegraphics[width=0.19\linewidth ]{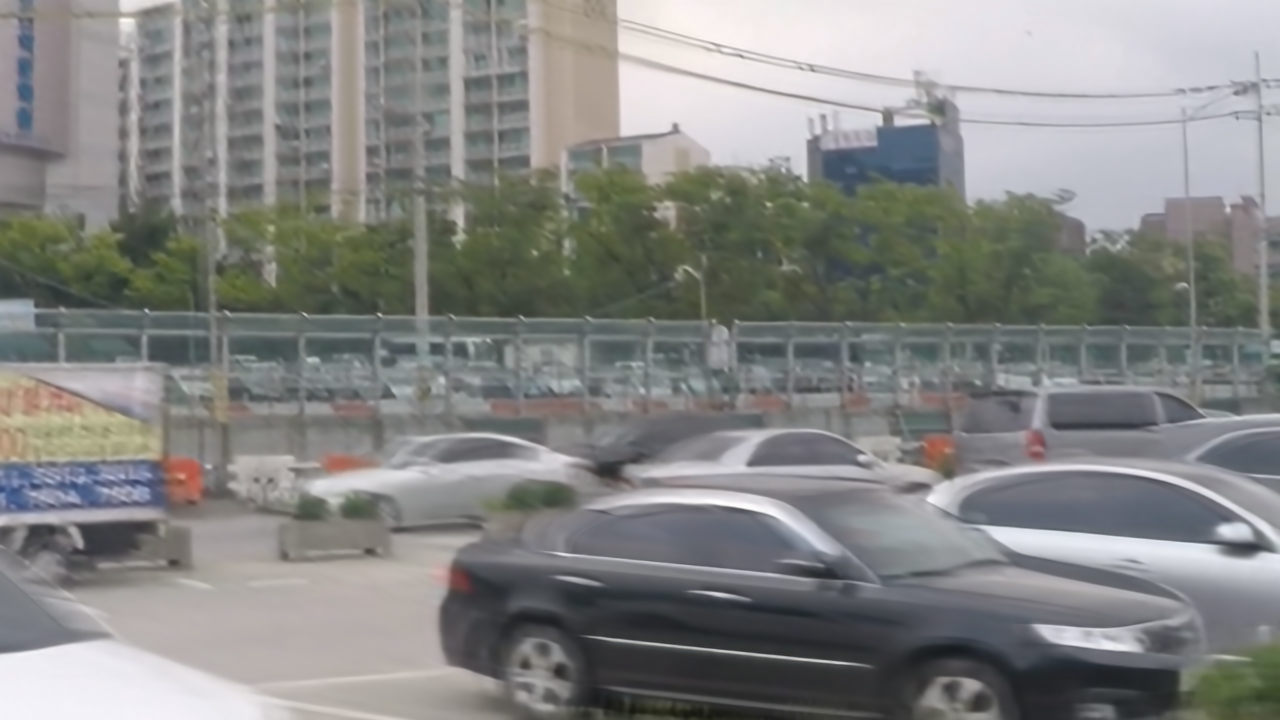}}
    \subfigure[]{
    \label{idea:g}
    \includegraphics[width=0.19\linewidth ]{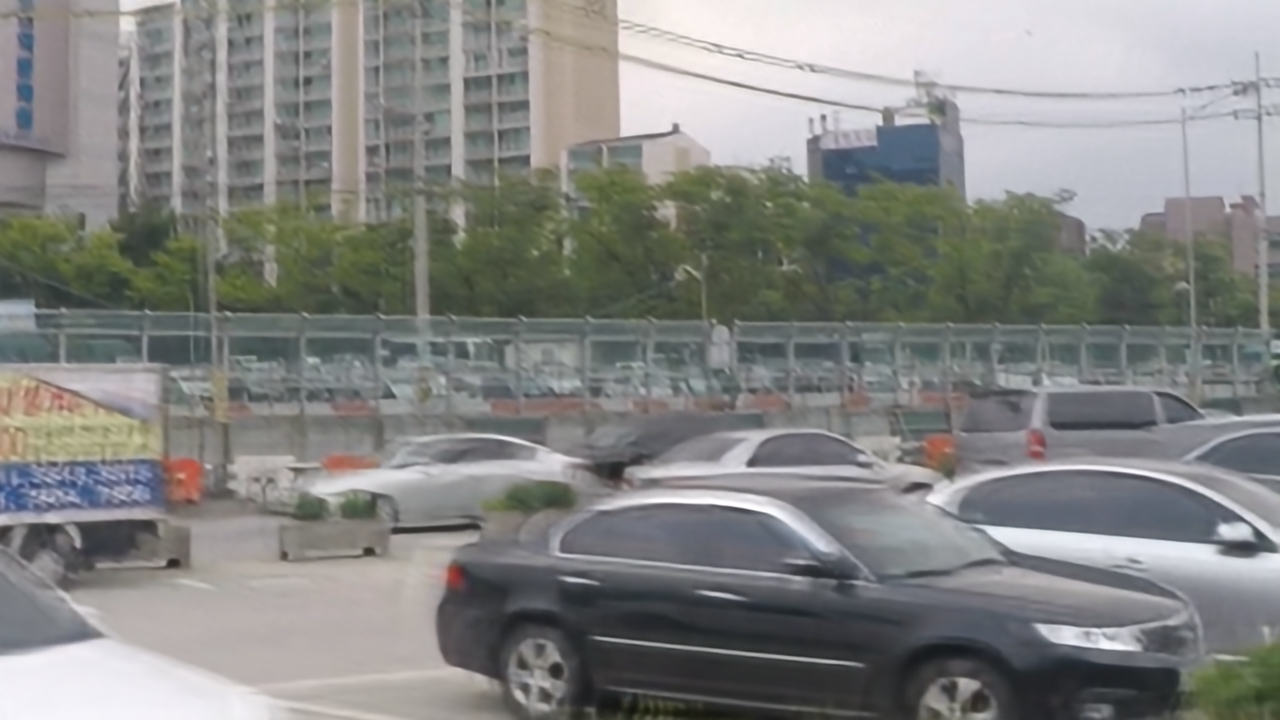}}
    \subfigure[]{
    \label{idea:h}
    \includegraphics[width=0.19\linewidth ]{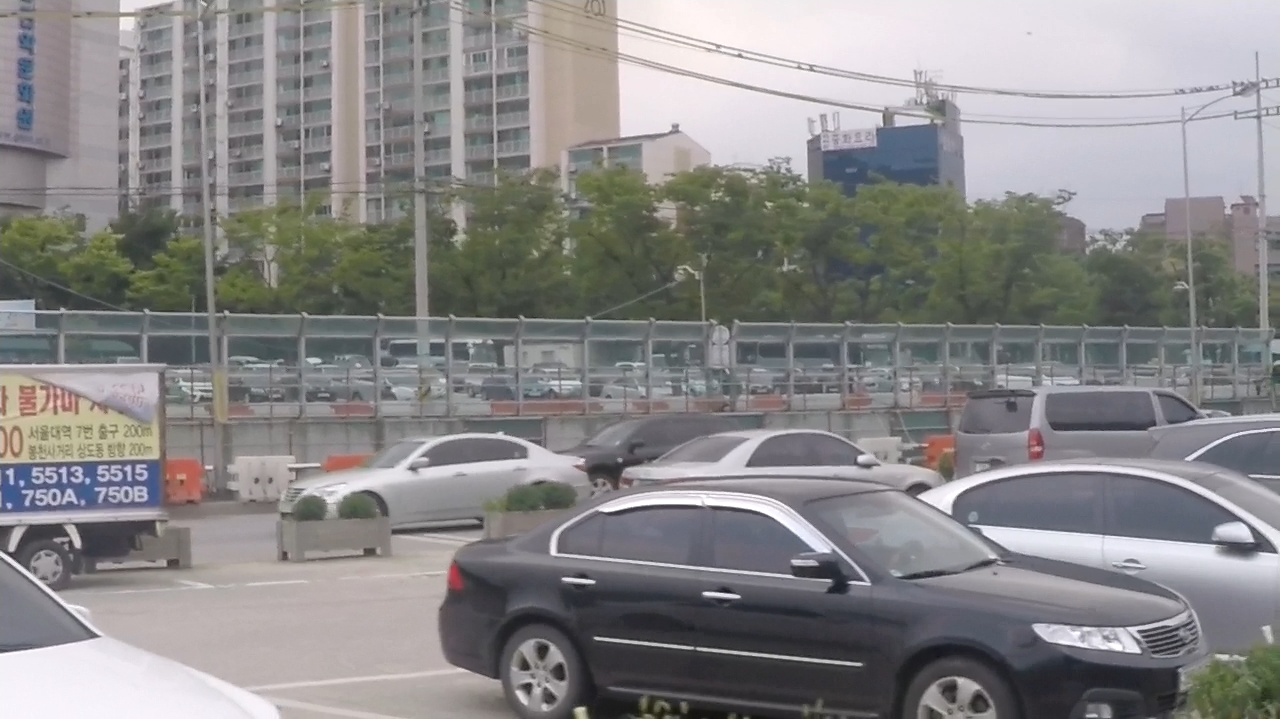}}
\caption{{\bf Comparison with state-of-the-art deblurring and super-resolution methods.} From left to right are input blurry image, RCAN+SRN, SRFBN+SRN, GFN and BMDSRNet, respectively. Zoom in the figure for better visibility.}
\label{figure_deblur_sr}
\end{figure*}

\begin{figure*}[t]
  \centering
  \subfigure[]{
    \label{idea:b}
    \includegraphics[width=0.19\linewidth ]{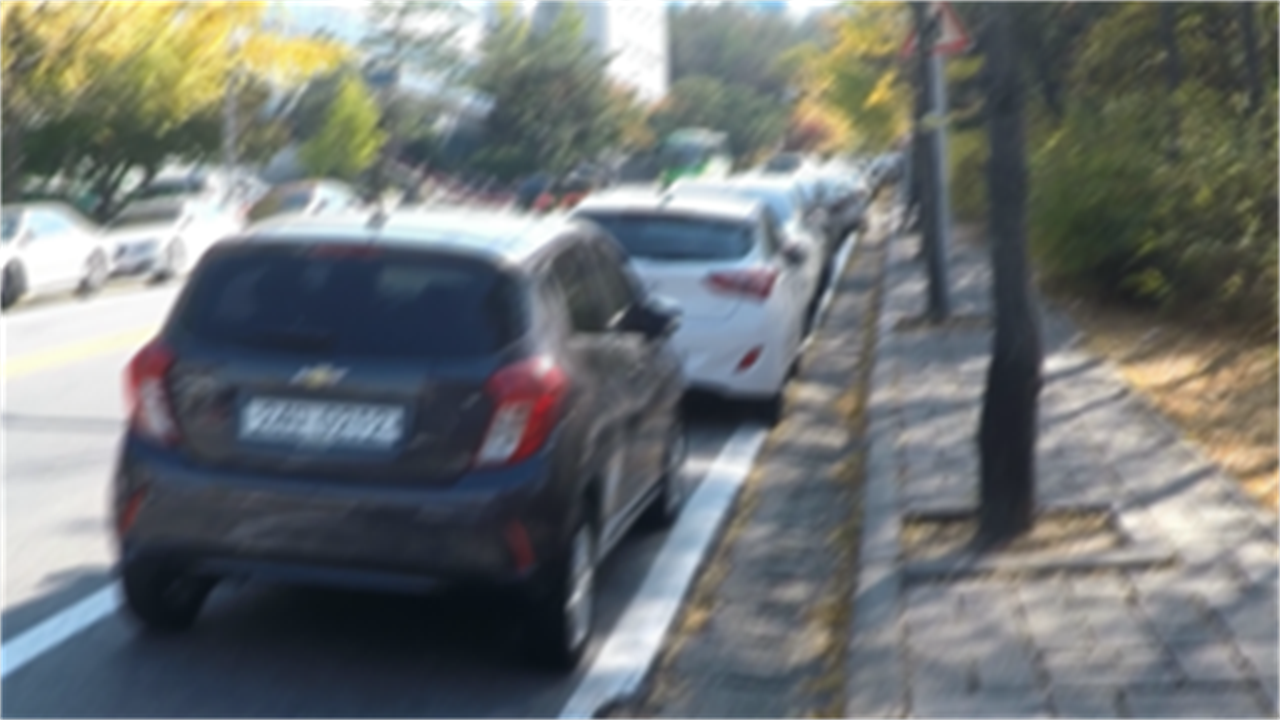}}
  \subfigure[]{
    \label{idea:d}
    \includegraphics[width=0.19\linewidth ]{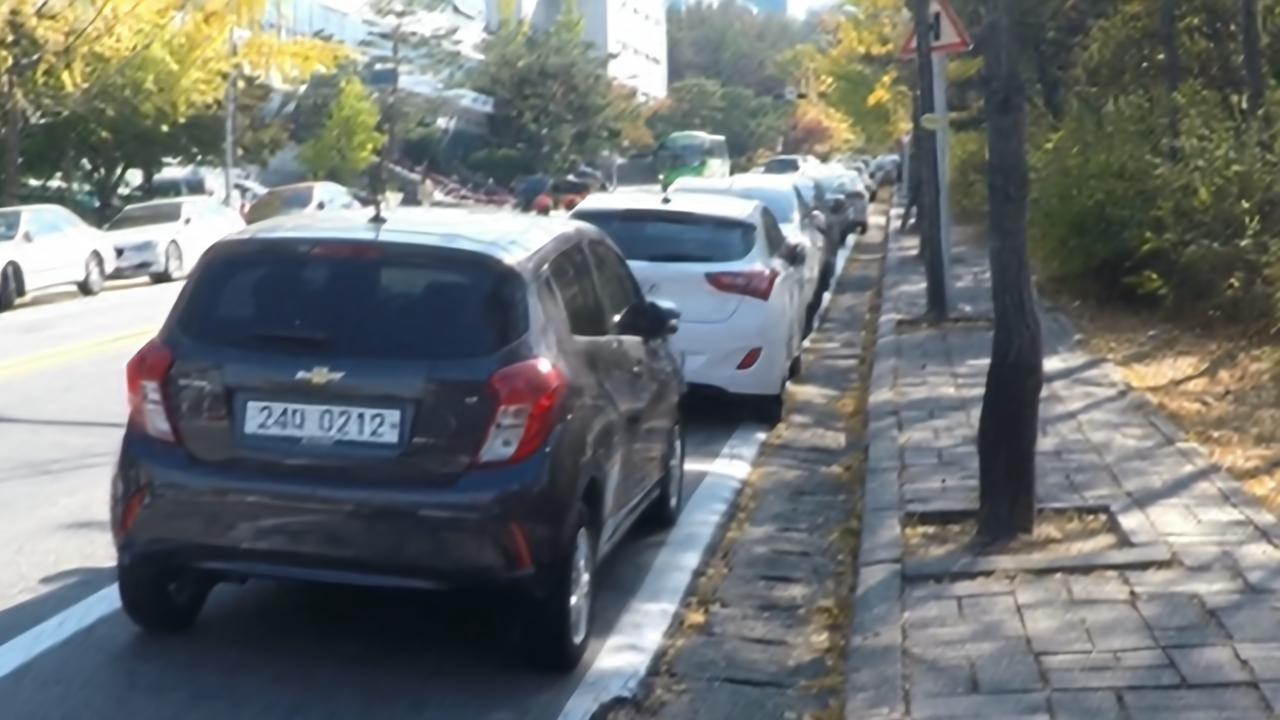}}
    \subfigure[]{
    \label{idea:f}
    \includegraphics[width=0.19\linewidth ]{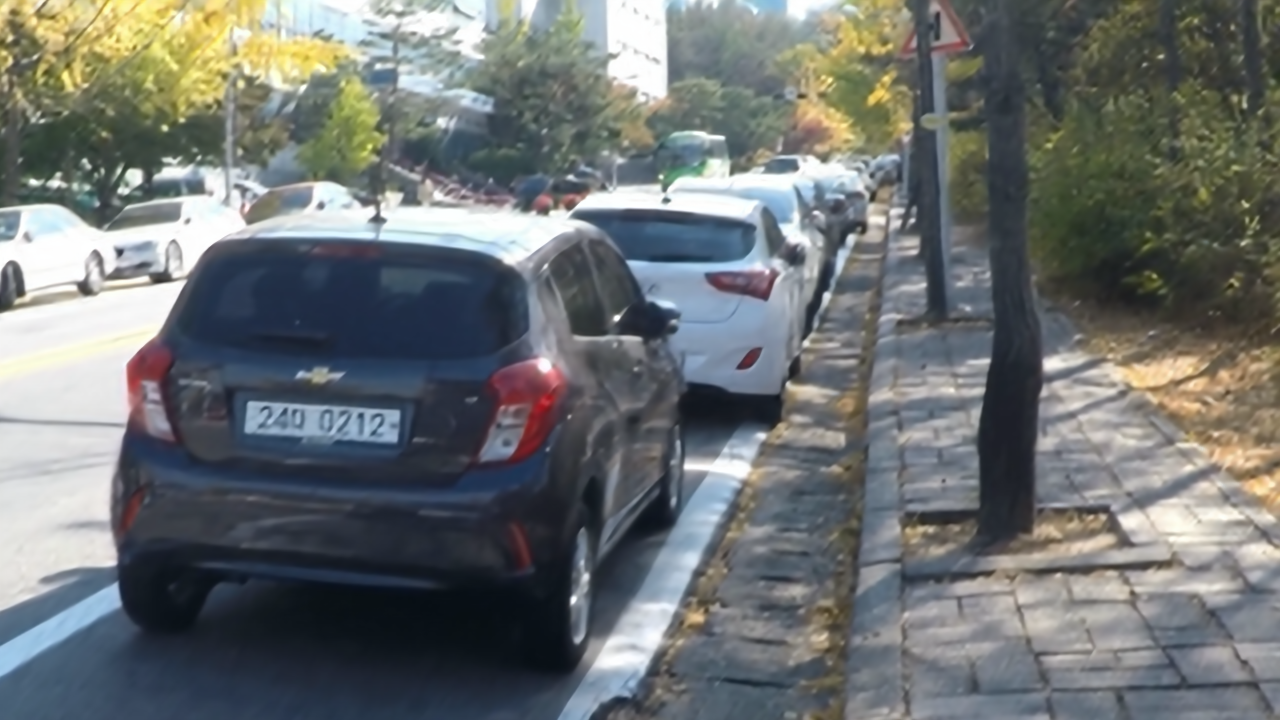}}
    \subfigure[]{
    \label{idea:g}
    \includegraphics[width=0.19\linewidth ]{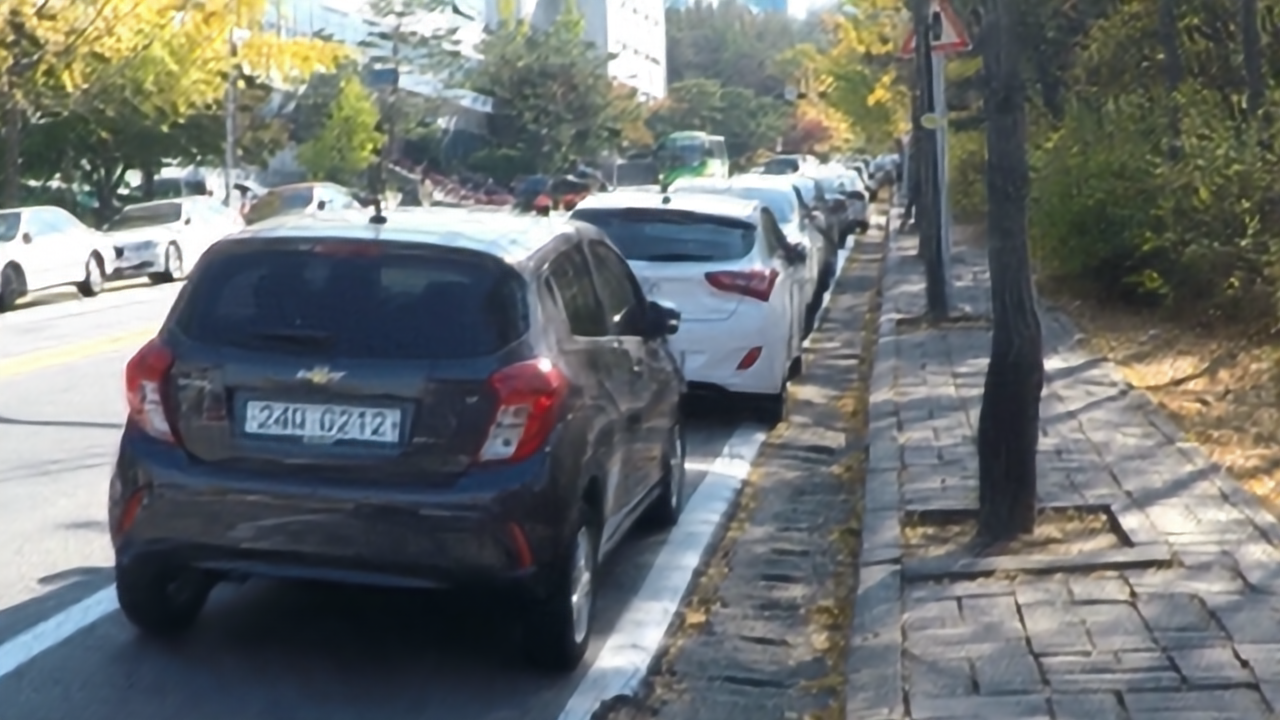}}
    \subfigure[]{
    \label{idea:h}
    \includegraphics[width=0.19\linewidth ]{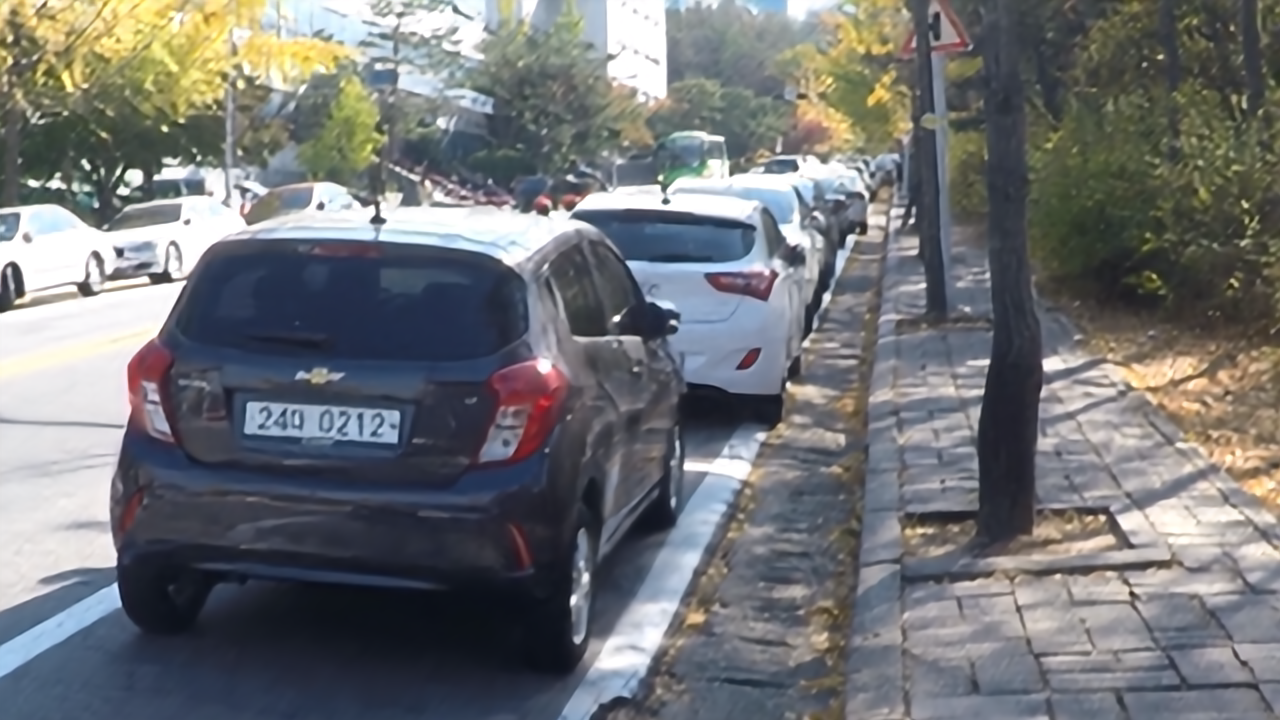}}
\caption{{\bf Comparison with state-of-the-art deblurring and super-resolution methods on the Motion-blurre LR GOPRO dataset \cite{nah2017deep}.} From left to right are the input blurry image, RCAN+SRN, SRFBN+SRN, GFN and BMDSRNet, respectively. Zoom in the figure for better visibility.}
\label{gopro_fig1}
\end{figure*}

\begin{figure*}[t]
  \centering
  \subfigure[]{
    \label{idea:b}
    \includegraphics[width=0.19\linewidth ]{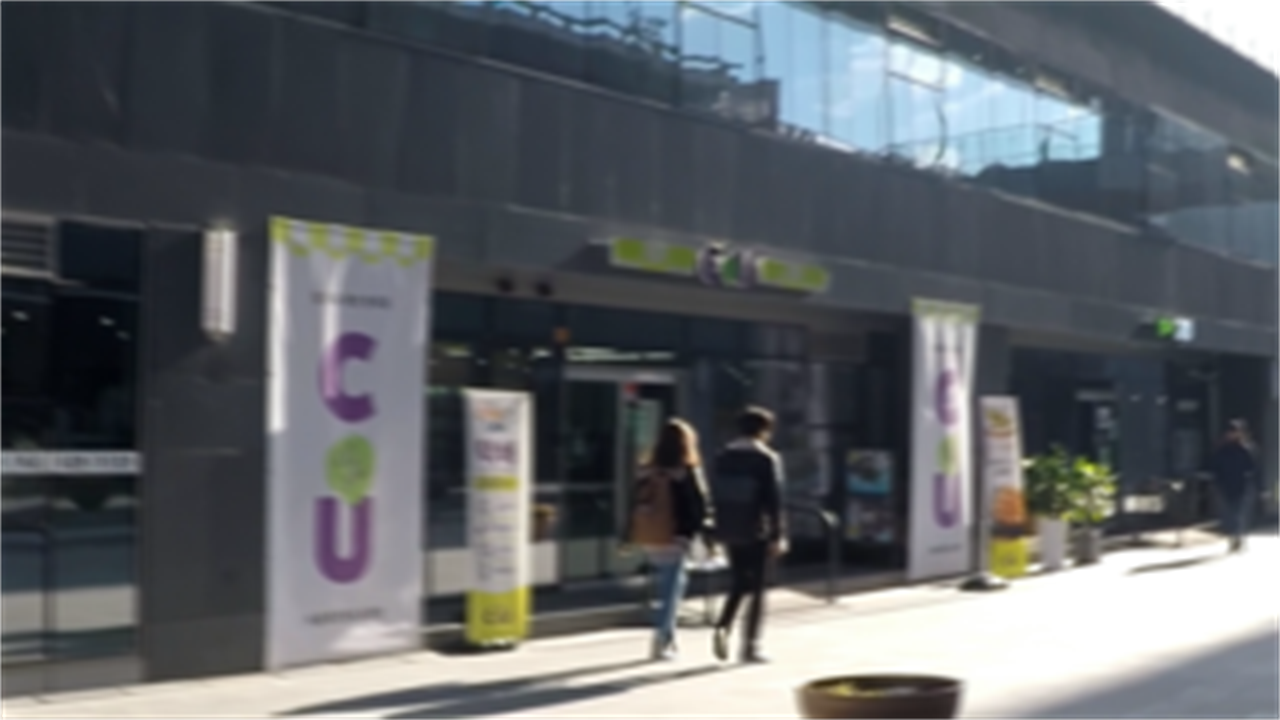}}
  \subfigure[]{
    \label{idea:d}
    \includegraphics[width=0.19\linewidth ]{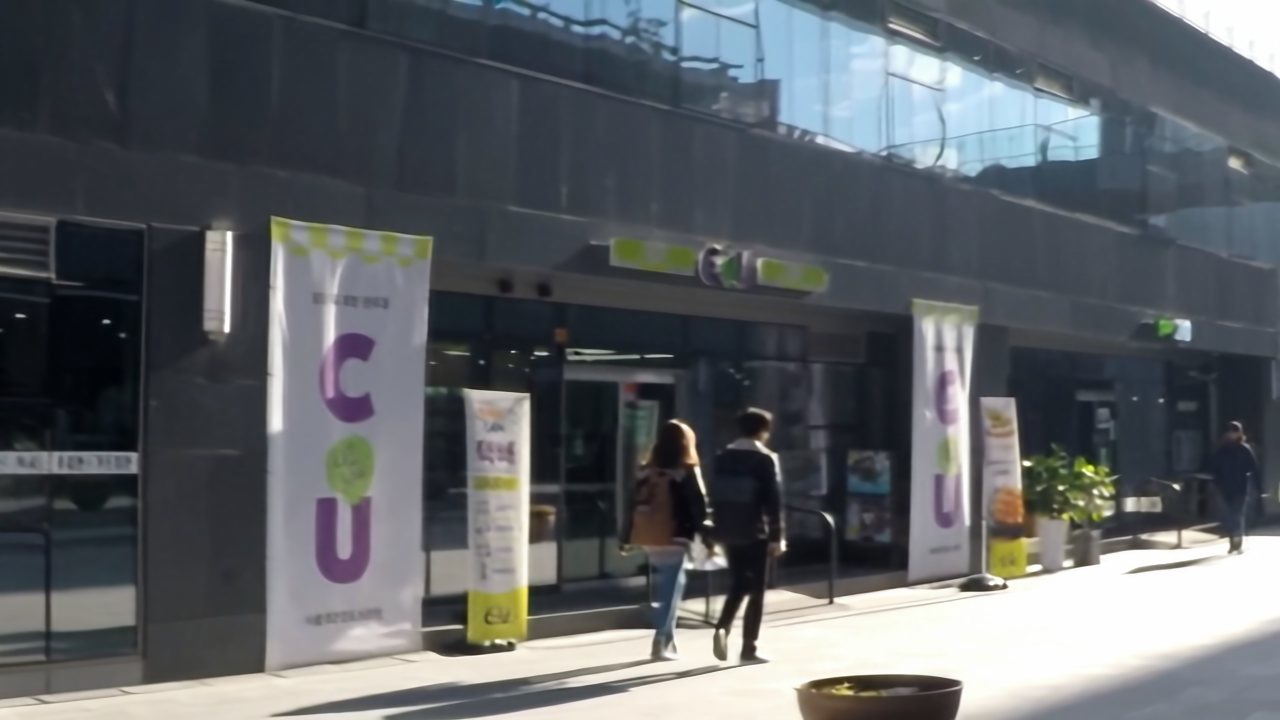}}
    \subfigure[]{
    \label{idea:f}
    \includegraphics[width=0.19\linewidth ]{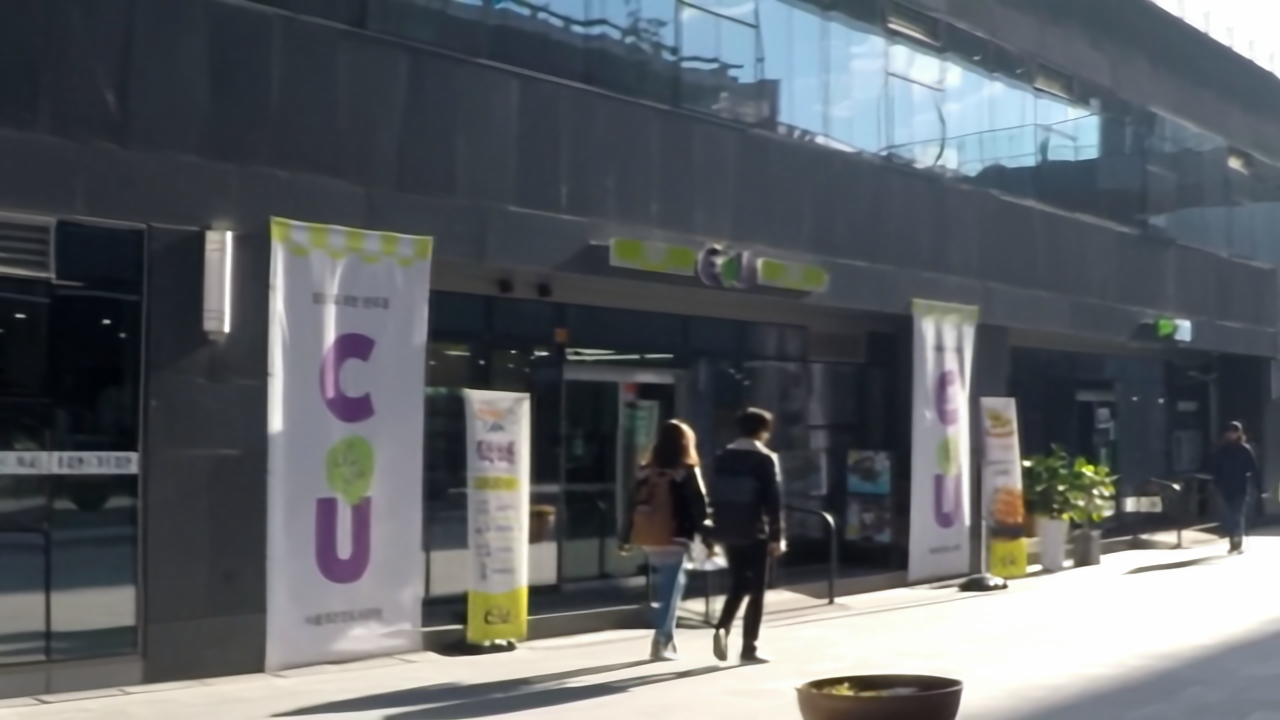}}
    \subfigure[]{
    \label{idea:g}
    \includegraphics[width=0.19\linewidth ]{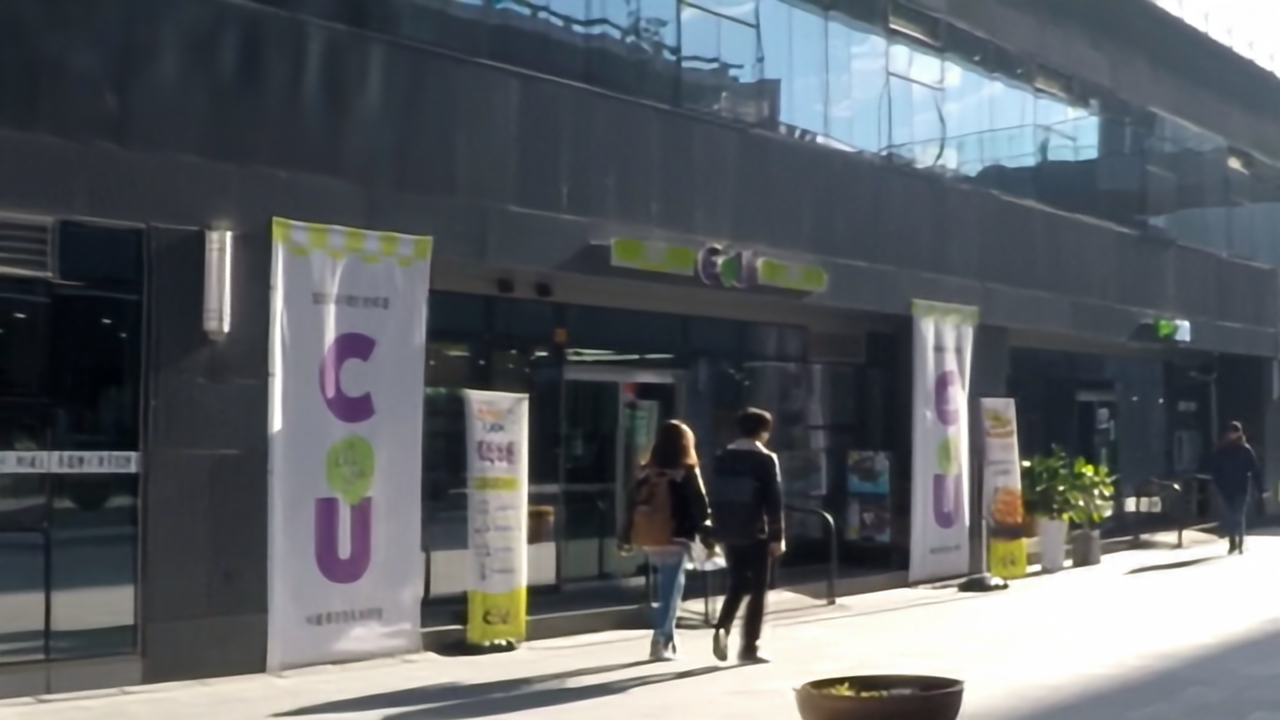}}
    \subfigure[]{
    \label{idea:h}
    \includegraphics[width=0.19\linewidth ]{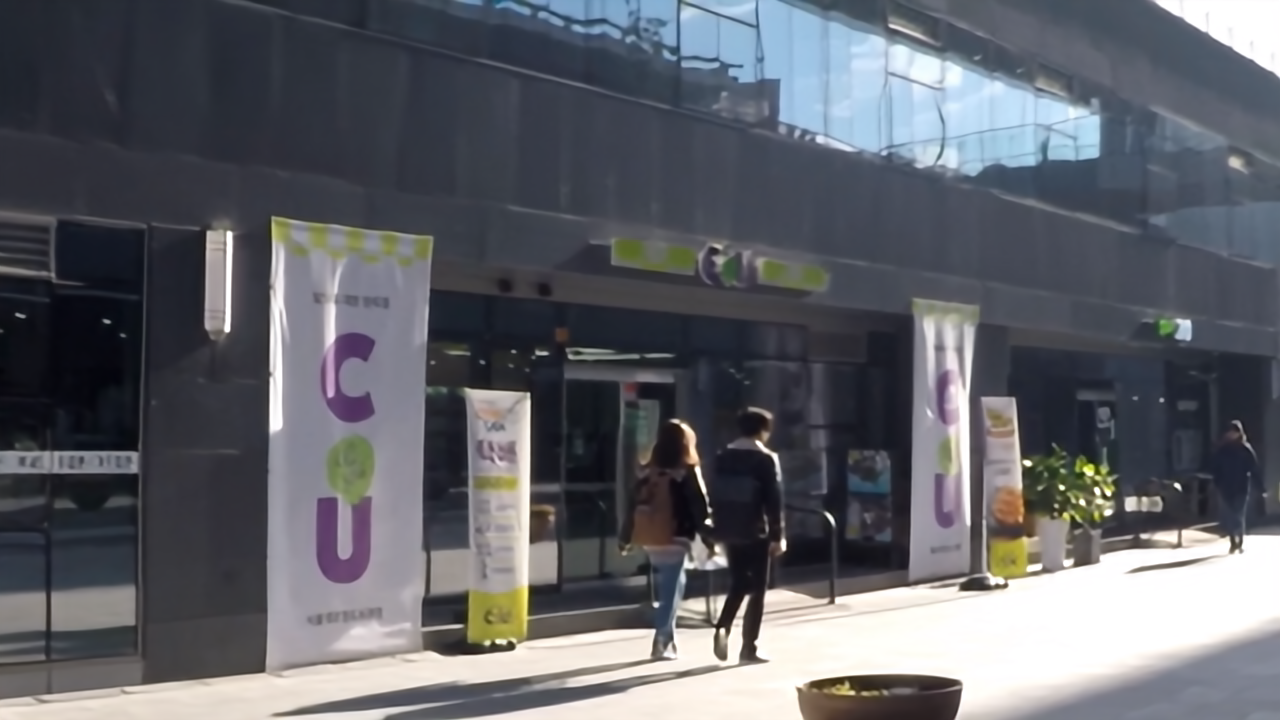}}
\caption{{\bf Comparison with state-of-the-art deblurring and super-resolution methods on the Motion-blurre LR GOPRO dataset \cite{nah2017deep}.} From left to right are the input blurry image, RCAN+SRN, SRFBN+SRN, GFN and BMDSRNet, respectively. Zoom in the figure for better visibility.}
\label{gopro_fig2}
\end{figure*}

\begin{figure*}[t]
  \centering
  \subfigure[]{
    \label{idea:a}
    \includegraphics[width=0.32\linewidth ]{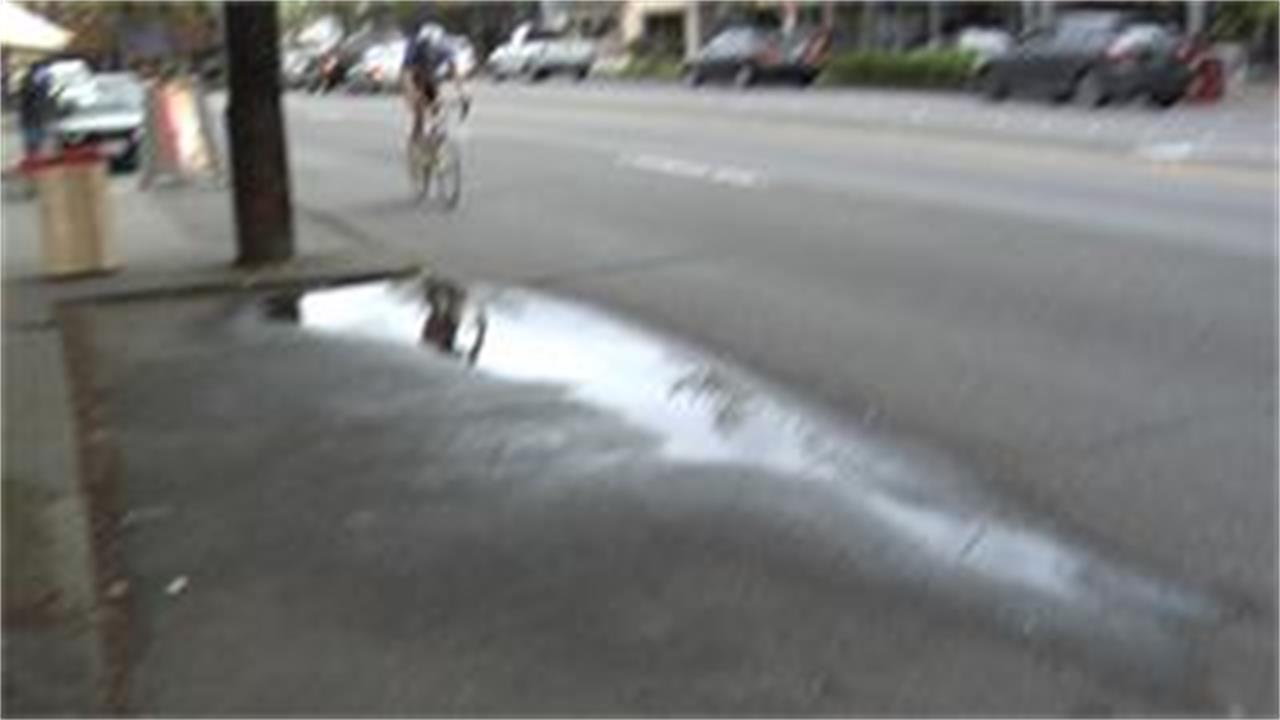}}
  \subfigure[]{
    \label{idea:b}
    \includegraphics[width=0.32\linewidth ]{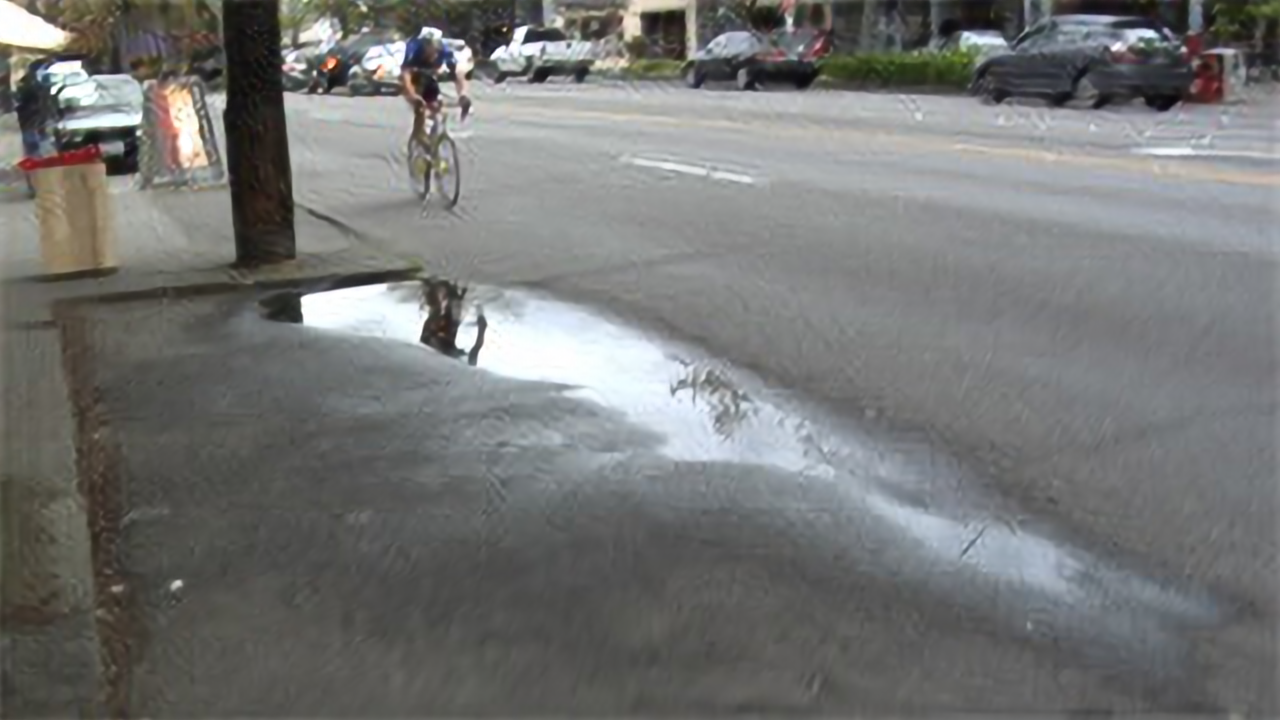}}
    \subfigure[]{
    \label{idea:c}
    \includegraphics[width=0.32\linewidth ]{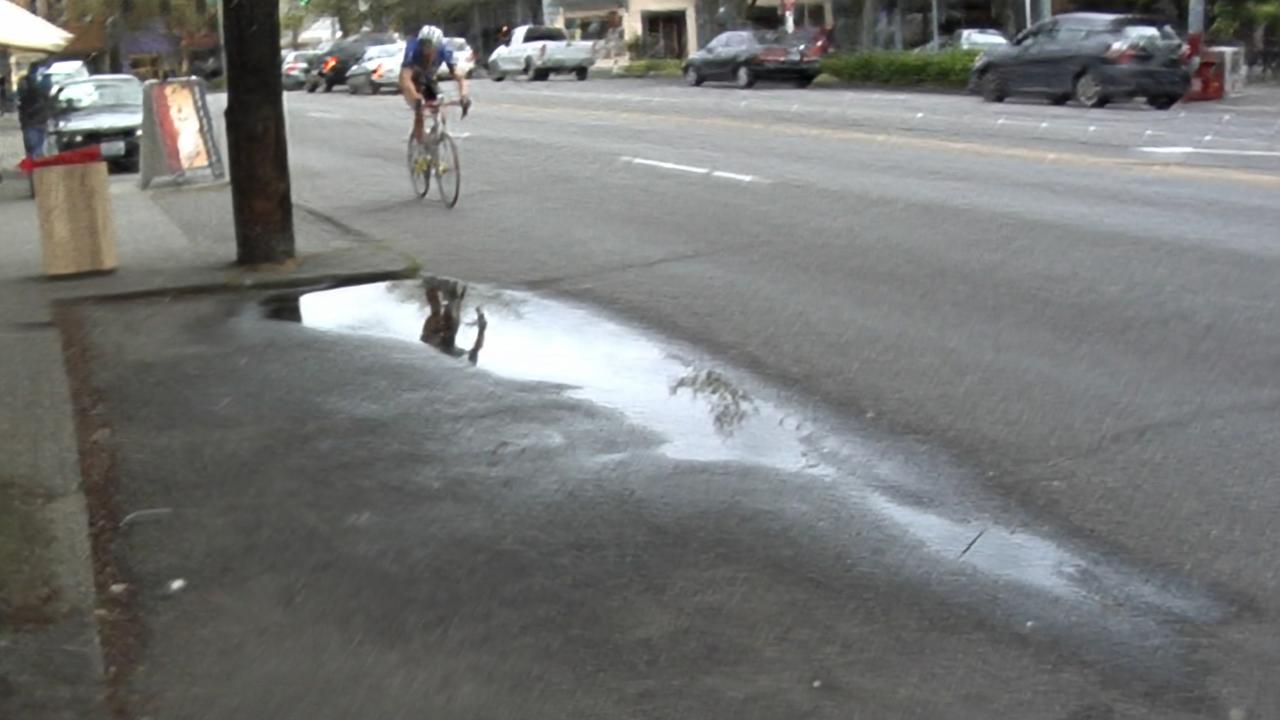}}
  \caption{{\bf Qualitative comparison on the real motion-blurred dataset \cite{su2017deep}}. \it The input image is shown in the first row. The second and third rows show results of the methods by GFN \cite{zhang2018gated} and BMDSRNet, respectively. Zoom in the figure for better visibility.}
  \label{real_world_fig1}
\end{figure*}

\subsection{Ablation Study}

In order to evaluate the effects of different components in the proposed model, we develop four networks, SRNet, BMDSRNet(C), BMDSRNet(F+C) and BMDSRNet(F+C+B). SRNet is a simplified version of BMDSRNet without the BMDNet module. BMDSRNet(C), BMDSRNet(F+C) and BMDSRNet(F+C+B) are three versions, whose inputs to the stage of \textit{LR to SR} are different. All these networks are illustrated in the following.  During the training stage, we update all weights with a mini-batch of size 4 in each iteration. 128 x 128 patches are cropped at random locations. The learning rate and training epoch is 0.0001 and 400, respectively. To train the final BMDSRNet, the weight of loss functions in “static to dynamtic” and “LR to SR” are the same.

\textbf{SRNet.} The architecture of SRNet is same to CoreNet. The main difference is that the input to SRNet is the blind motion-blurred images, rather than the deblurred results from BMDNet.

\textbf{BMDSRNet(C).} It consists of BMDNet and CoreNet. One motion-blurred image is put into BMDNet to recover seven deblurred low-resolution images. Then we input the central frame of them to CoreNet to generate the sharp high-resolution image.

\textbf{BMDSRNet(F+C).} It consists of BMDNet, CoreNet, ForNet and FuNet. The main difference from BMDSRNet(C) is that it uses an additional ForNet module to extract temporal information from the deblurred neighboring frames. Then the FuNet generates the final deblurred high-resolution images based on the results form CoreNet and ForNet.

\textbf{BMDSRNet(F+C+B).} This architecture is our whole blind motion-deblurred super-resolution networks. It consists of BMDNet, CoreNet, ForNet, BackNet and FuNet. ForNet and BackNet extract temporal information from bidirection to help FuNet generate final deblurred high-resolution images.

\textbf{Results on different scales.} We report the PSNR and SSIM of the aforementioned four models on three down-sampling scales. The quantitative and visual results are shown in Tab. \ref{table_ablation} and Fig. \ref{figure_ablation_study}. The performance difference between BMDSRNet(C) and SRNet illustrates the effect of our ``divide-and-conquer"" strategy. BMDSRNet(F+C) achieves better performance than BMDSRNet(C), which corresponds to the well-known knowledge that multi-frame super-resolution methods have advantage over the single image super-resolution methods. This demonstrates the effect of the \textit{``static-to-dynamic"} stage. The BMDSRNet(F+C+B) is better than BMDSRNet(F+C), suggesting extracting temporal features from bidirection is helpful to generate high-resolution images.

\begin{figure*}[t]
  \centering
  \subfigure[]{
    \label{idea:a}
    \includegraphics[width=0.32\linewidth ]{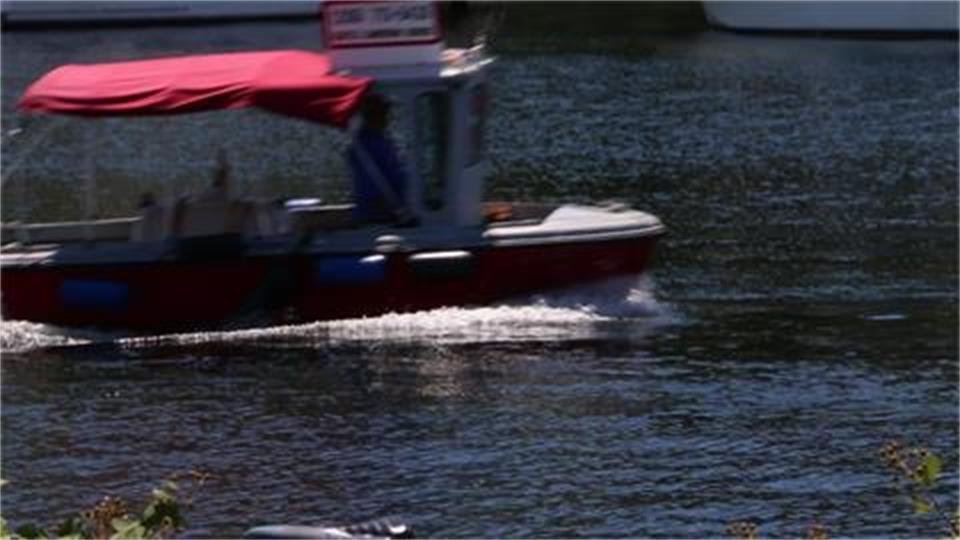}}
  \subfigure[]{
    \label{idea:b}
    \includegraphics[width=0.32\linewidth ]{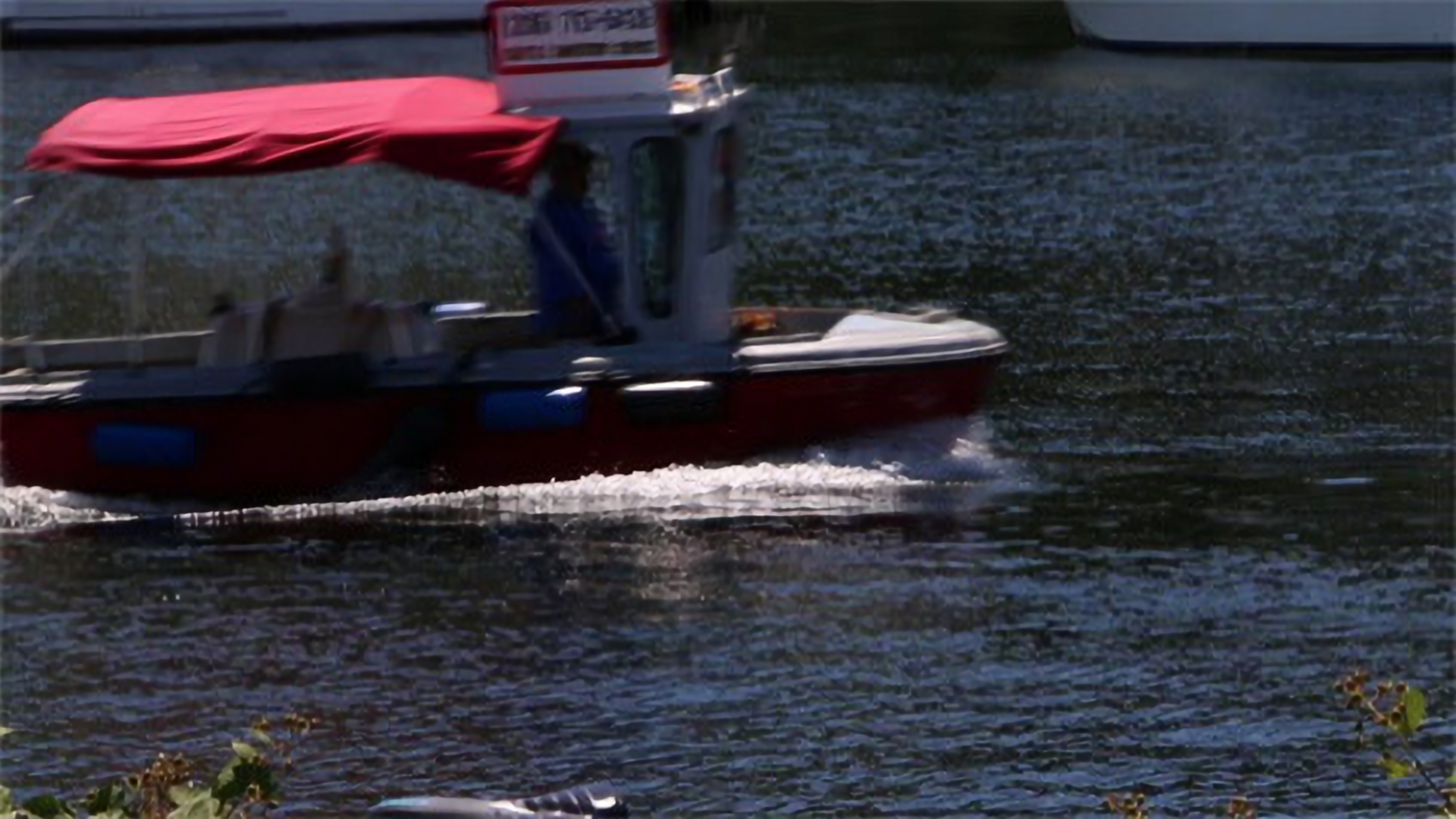}}
    \subfigure[]{
    \label{idea:c}
    \includegraphics[width=0.32\linewidth ]{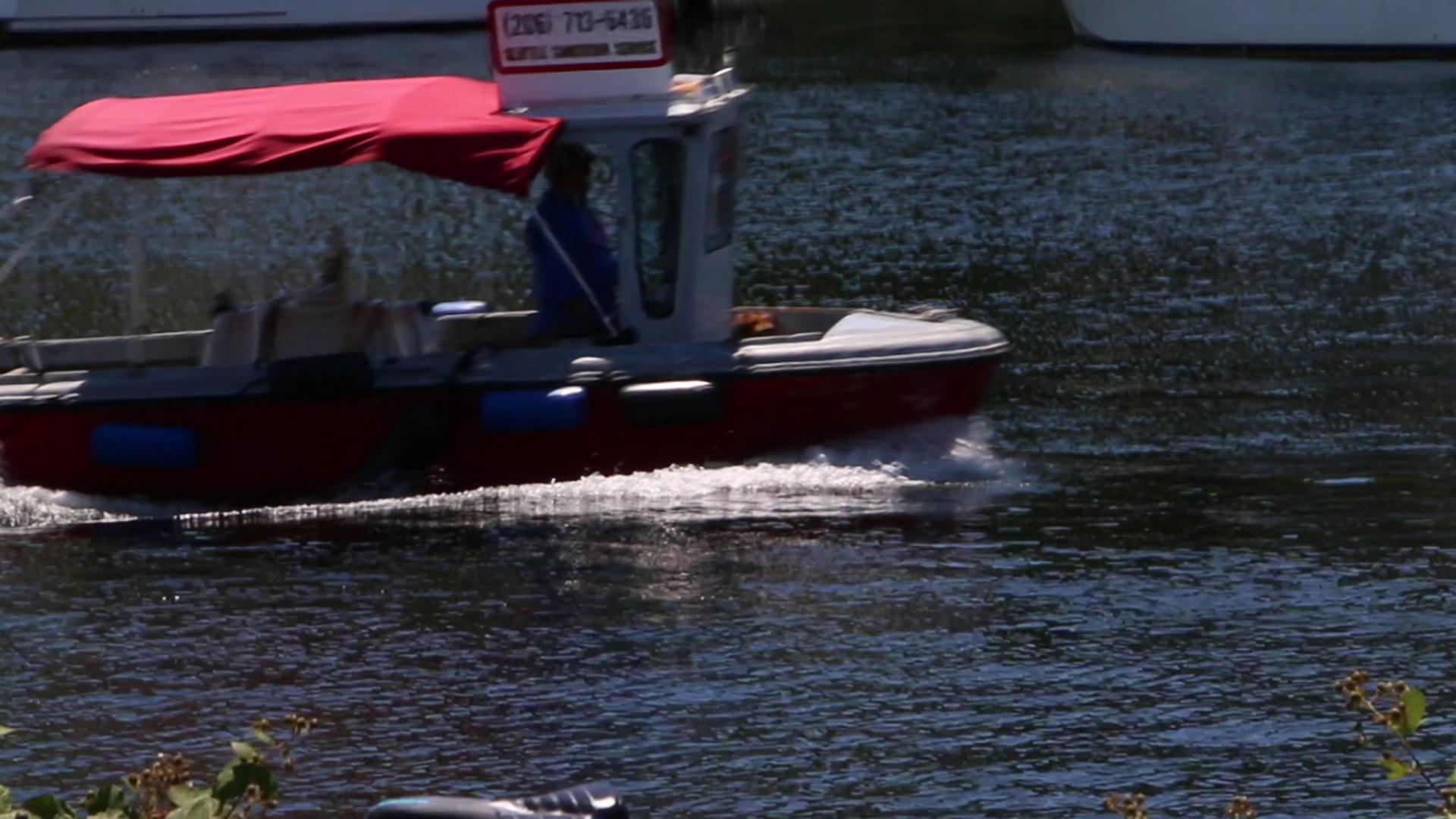}}
  \caption{{\bf Qualitative comparison on the real motion-blurred dataset \cite{su2017deep}}. \it The input image is shown in the first row. The second and third rows show results of the methods by GFN \cite{zhang2018gated} and BMDSRNet, respectively. Zoom in the figure for better visibility.}
  \label{real_world_fig2}
\end{figure*}

\begin{figure*}[t]
  \centering
  \subfigure[]{
    \label{idea:a}
    \includegraphics[width=0.32\linewidth ]{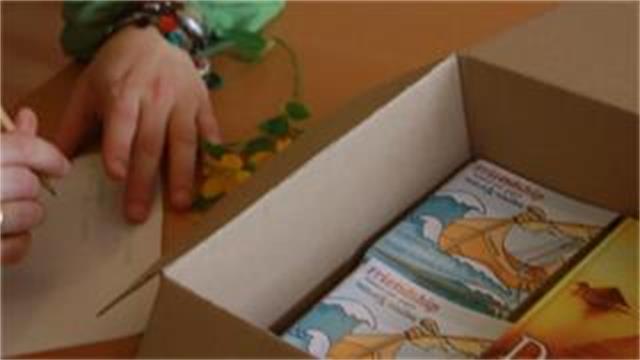}}
  \subfigure[]{
    \label{idea:b}
    \includegraphics[width=0.32\linewidth ]{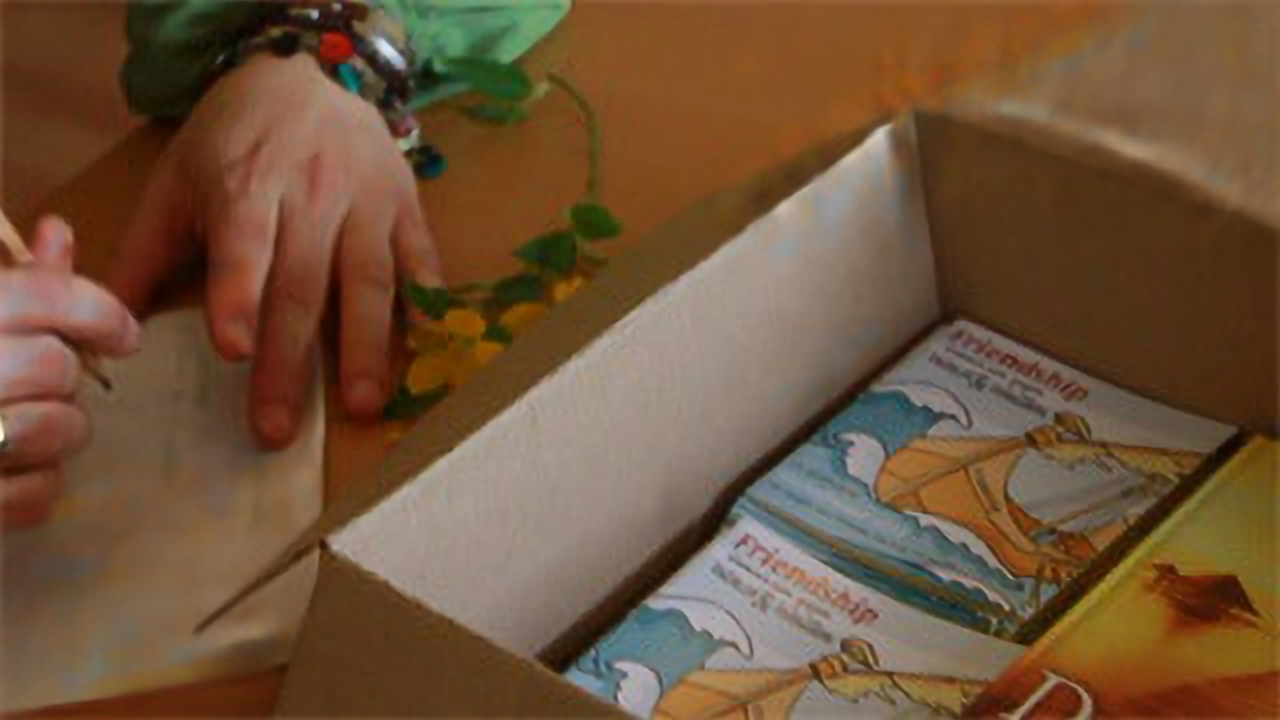}}
    \subfigure[]{
    \label{idea:c}
    \includegraphics[width=0.32\linewidth ]{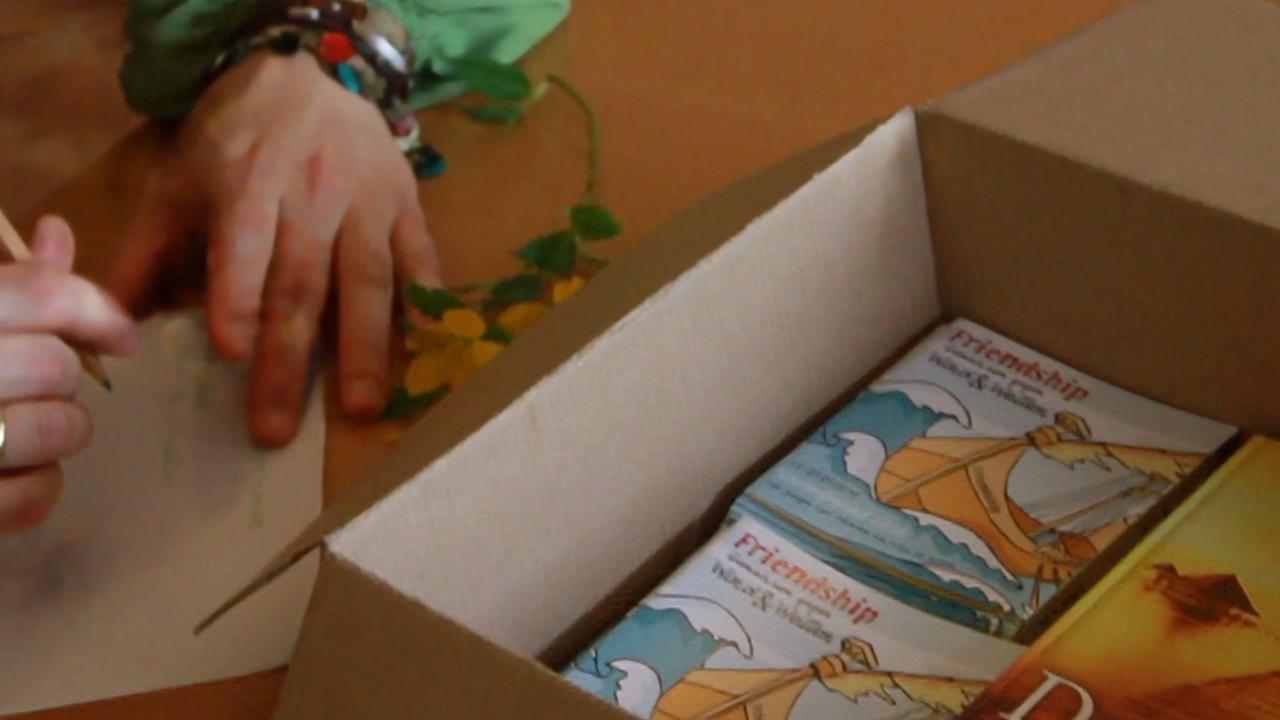}}
  \caption{{\bf Qualitative comparison on the real motion-blurred dataset \cite{su2017deep}}. \it The input image is shown in the first row. The second and third rows show results of the methods by GFN \cite{zhang2018gated} and BMDSRNet, respectively. Zoom in the figure for better visibility.}
  \label{real_world_fig3}
\end{figure*}

\begin{table*}[!tb]
  \centering 
    \caption{Performance of different model structures on the Motion-Blurred LR dataset in terms of PSNR and SSIM for scale factors 2, 3 and 4, respectively.}
    \setlength\tabcolsep{5.0pt}
    \begin{tabular}{l | c | c | c | c | c | c }
    \toprule
    \multirow{2}{*}{Method} & \multicolumn{2}{c|}{$\times 2$}&\multicolumn{2}{c|}{$\times 3$} &\multicolumn{2}{c}{$\times 4$}\\\cline{2-7}
    
   & PSNR &  SSIM & PSNR & SSIM & PSNR & SSIM \\
    \hline
    RCAN & 27.57 & 0.8908 & 27.18 & 0.8846 & 26.65 & 0.8737 \\
    RCAN + SRN & 30.94 & 0.9443 & 29.54 & 0.9264 & 28.18 & 0.9029  \\
    SRN + RCAN & 28.88 & 0.9180 & 27.08 & 0.9059 & 27.56 & 0.8938 \\
    \hline
    SRFBN & 27.57 & 0.8907 & 27.16 & 0.8843 & 26.63 & 0.8730  \\
    SRFBN + SRN & 30.92 & 0.9196 &29.50  & 0.8901  & 28.09 & 0.8504  \\
    SRN + SRFBN & 30.40 & 0.9119 & 29.03 & 0.8856 & 27.63 & 0.8400 \\
    \hline
    GFN & - & - & - & - & 27.47 & 0.8926 \\
    \textbf{BMDSRNet} & \textbf{31.62} & \textbf{0.9483} & \textbf{30.44} & \textbf{0.9356} & \textbf{28.78} & \textbf{0.9132} \\
    \bottomrule
    \end{tabular}%
    \label{table_deblur_sr1}
\end{table*}%

\subsection{Comparison with State-of-the-art Methods}

We have verified the effects of different parts of the proposed BMDSRNet. In this section, we compare our method with other state-of-the-art algorithms, including two SR methods (RCAN \cite{zhang2018image}, SRFBN \cite{li2019feedback}), one motion-deblurred method (SRN \cite{tao2018scale}), and methods which jointly address SR and image deblurring tasks. RCAN is one of the best methods for bicubic degradation. SRFBN consists of several feedback blocks and achieves state-of-the-art performance for SR. 
SRN is one popular image deblurring method. Specially,it is trained without adversarial loss and one of the state-of-the-art methods on removing motion blur \cite{rim2020real}. 
We directly use the official deblurred models which are trained on the GOPRO dataset. For fair comparisons, we also re-train the two SR methods on the GOPRO dataset.
The values of PSNR and SSIM of different methods on three different motion-blurry LR sets are shown in Tab. \ref{table_deblur_sr1} and Fig. \ref{figure_deblur_sr}. Results show that the proposed BMDSRNet outperforms two popular SR methods, and the combination of SR and image deblurring methods, and the previous blind deblurring super-resolution network, \textit{i.e.}, GFN.

\subsection{Test on the LR GOPRO Dataset}

In order to further evaluate the proposed method, we also test it on the LR GOPRO dataset. This dataset is synthesized based on GOPRO dataset via down-sampling to downscale blurry images by $4 \times$ to generate blurry LR images. The ground-truth sharp high-resolution images are the original sharp images in the GOPRO dataset.

We compare our method with two SOTA SR methods (SRResNet \cite{ledig2017photo}, EDSR \cite{lim2017enhanced}), two joint image deblurring and SR approaches ( SCGAN \cite{xu2017learning}, GFN \cite{zhang2018gated}), and the combinations of blind image deblurring algorithms (DeepDeblur \cite{nah2017deep}) and SR algorithms (SRResNet \cite{ledig2017photo}, ED-DSRN\cite{zhang2018deep}). 
The values of PSNR and SSIM of different methods on the LR GOPRO set are shown in Tab. \ref{table_deblur_sr2}. Fig. \ref{gopro_fig1} and \ref{gopro_fig2} show the qualitative results. The results show that the proposed BMDSRNet does not only outperform the traditional SR methods, and the joint image deblurring and SR approaches, but also achieves better performance than the previous state-of-the-art method, GFN, which is also designed for blind motion deblurring super-resolution.

As we all know that blind motion deblurring super-resolution is a more ill-posed problem. The proposed BMDSRNet achieves better performance because we employ the “divide-and-conquer” scheme, rather than in-one-go SR network. Results are reported in Tab. \ref{table_deblur_sr2}. Firstly, the RCAN and SRFBN are two state-of-the-art SR methods, which can directly super-resolve an LR blurry image to an HR sharp image. However, these methods achieve worse performance than RCAN + SRN and SRFBN + SRN. SRN is one of the state-of-the-art deblurring methods. It shows that “divide-and-conquer” scheme is a better option than an in-one-go network for the blind motion deblurring super-resolution. Secondly, the SRFBN + SRN and SRFBN + SRN achieve worse performance than the proposed BMDSRNet, showing that extracting a sequence from a blurry image is better than using only the intermediate sharp image. Thirdly, the previous work \cite{jin2018learning} hows that the intermediate sharp image in the extracted sharp sequence is of high-quality with satisfied PSNR values, guaranteeing the quality of intermediate sharp images. Finally, SRN + RCAN and SRN + SRFBN, which firstly predict one sharp LR image and then super-resolve to corresponding HR version, achieve worse performance than firstly extracting a sharp sequence of LR images. It verifies that “only predict one sharp LR image” is worse than predicting a sharp sequence.

\begin{table}[!tb]
  \centering 
    \caption{Performance of different model structures on the LR GORPO dataset in terms of PSNR and SSIM for scale factor 4.}
    \setlength\tabcolsep{5.0pt}
    \begin{tabular}{l | c c c c c c c c }
    \toprule
    Method & PSNR &  SSIM & Params & Times \\
    \hline
    SCGAN & 22.74 & 0.783 & 1.1M & 0.66  \\
    SRResNet &  24.40 &  0.827 & 1.5M & 0.07 \\
    EDSR & 24.52 &   0.836 & 43M & 2.10\\
    \hline
    DeepDeblur + SRResNet & 24.99 &  0.827 & 13M & 0.66 \\
    SRResNet + DeepDeblur & 25.93 & 0.850  & 13M & 6.06 \\
    \hline
    DeepDeblur + ED-DSRN & 21.53  &  0.682 & 54M & 2.18 \\
    ED-DSRN + DeepDeblur &  24.66 &  0.827 & 54M & 2.95  \\
    \hline
    GFN & 27.74 &  0.896 & 12M & 0.07\\ 
    \textbf{BMDSRNet} & \textbf{28.15}  & \textbf{0.904} & \textbf{2.9M} & \textbf{0.11} \\
    \bottomrule
    \end{tabular}%
    \label{table_deblur_sr2}
\end{table}%

\subsection{Test on the real motion-blurred Dataset}

Then we compare the performance of our approach with the previous state-of-the-art blind motion-blurred SR method, \textit{i.e.}, GFN \cite{zhang2018gated}, on real blurry images \cite{su2017deep}. The results can refer to Fig. \ref{real_world_fig1}, \ref{real_world_fig2} and \ref{real_world_fig3}.

\section{Conclusion}

In this paper, we propose a blind motion deblurring super-resolution networks to recover sharp high-resolution images from motion-blurred low-resolution input. Our main contribution is the novel use of spatio-temporal information implied in motion-blurred images for single image super-resolution. We first design a motion deblurring network which can model the reverse process of generation of motion-blurred images. This network can extract several neighbouring frames and thus transfer the static super-resolution to dynamic super-resolution. Then a well-designed three-streams network is developed to learn bidirectional spatio-temporal information to recover the final sharp high-resolution images. The experimental results on two datasets suggest that the proposed model outperforms existing methods for super-resolving blind motion-blurred images.

\section*{Acknowledgment}
This work is funded in part by the ARC Centre of Excellence for Robotics Vision (CE140100016),  ARC-Discovery (DP 190102261) and ARC-LIEF (190100080) grants, as well as a research grant from Baidu on autonomous driving.  The authors gratefully acknowledge the GPUs donated by NVIDIA Corporation. We thank all anonymous reviewers and editors for their constructive comments.

\bibliographystyle{IEEEtran}
\bibliography{egbib}
%




\end{document}